\documentclass{article}

\usepackage[final]{corl_2026} 

\usepackage{amsmath}
\usepackage{amssymb}
\usepackage{booktabs}
\usepackage{multirow}
\usepackage{array}
\usepackage{xspace}
\usepackage{enumitem}
\usepackage{graphicx}
\usepackage{wrapfig}
\usepackage{capt-of}
\usepackage{url}
\usepackage{tikz}
\usetikzlibrary{arrows.meta,calc,positioning}
\usepackage{pgfplots}
\pgfplotsset{compat=1.18}

\newcommand{\system}{RoboLineage\xspace}
\newcommand{\artifact}[1]{\texttt{#1}}
\newcommand{\projectpage}{\url{https://robolineage.github.io/}}

\title{\system: Agent-Native Data Lifecycle Governance Across Robot Policy Iterations}

\author{
\begin{tabular}{c}
\vspace{1.4em}\\[-0.6em]
\normalsize
Qian Luo\textsuperscript{1,2} \quad
Wentao Guo\textsuperscript{1,3} \quad
Zhennan Qin\textsuperscript{1} \quad
Nanchun Guo\textsuperscript{1}
\\[-0.1em]
\normalsize
Yunhan Zhao\textsuperscript{1} \quad
Yi Ma\textsuperscript{1,2} \quad
Yanchao Yang\textsuperscript{1,2}
\\[0.35em]
\normalfont\small
\textsuperscript{1}TranscEngram \quad
\textsuperscript{2}The University of Hong Kong \quad
\textsuperscript{3}Beijing Institute of Technology
\end{tabular}
}

\begin{document}
\maketitle

\begin{center}
\vspace{-1.2em}
\small Project page: \projectpage
\vspace{1.2em}
\end{center}


\begin{abstract}
We present RoboLineage, an agent-native data lifecycle governance system for
robot policy iteration. Modern robot policies improve through repeated cycles
of data collection and retraining. Teams still rely on local tools, scripts,
and expert memory to review the evidence and decide which data to use or which
policy to release. \system records how rollout evidence leads to dataset changes
and how those datasets produce policy versions. Evaluation and release decisions
remain linked to that history, including the plan for the next collection round.
Agents interpret embodied evidence and prepare accepted data for existing
training stacks. They also assess dataset readiness and carry context across
iterations through typed, validated artifacts. In real-robot
manipulation workflows, \system makes routine policy iteration faster and more
auditable while maintaining downstream policy performance. We open source
\system as a lightweight lifecycle layer for different robot embodiments and
training families.
\end{abstract}



\section{Introduction}
\label{sec:introduction}

Robot learning has become increasingly data-rich. Modern imitation learning and
vision-language-action (VLA) pipelines can train capable policies from
teleoperated demonstrations, in-the-wild datasets, and policy-specific
corrections
~\citep{zhao2023aloha,chi2023diffusion,brohan2022rt1,brohan2023rt2,oneill2023openx,khazatsky2024droid,kim2024openvla}.
At the same time, the day-to-day work of improving a real robot policy is still
surprisingly manual. After collecting a rollout batch, the team reviews the
evidence and curates a training set. It then prepares inputs for the learner and
retrains the policy. Evaluation determines whether to release the new version or
collect more data. These steps form the data lifecycle through which a robot
policy improves.

Expert robotics groups already manage this lifecycle with care. Their
established procedures cover data preparation through policy release.
The difficulty is that these practices rarely become a durable
accountability chain across policy iterations. Human experts repeatedly
reconstruct which execution evidence justified a dataset change, which data
trained a checkpoint, why an evaluation was trusted, and what should be
collected next. This reconstruction is fragile: evidence is scattered, decisions
are remembered through local practice, and lineage across iterations remains
weak until experts piece it back together. This fragility matters more as
datasets span multiple robots and training moves toward generalist policies.
Recent work shows that policy quality depends on which data is selected and
how it is reused
~\citep{dasari2019robonet,fang2023rh20t,oneill2023openx,lin2024datascaling,octo2024}.

Robot policy iteration calls for explicit lifecycle governance. Rollouts,
datasets, and trained policies need links to the observations, reviews,
admission rules, and evaluations that justify their use. Recording these links
across iterations makes it possible to audit a dataset change, trace a
checkpoint to its training inputs, and recover the evidence behind a deployment
decision.

\begin{wrapfigure}{R}{0.55\linewidth}
    \vspace{-0.8em}
    \centering
    \includegraphics[width=\linewidth]{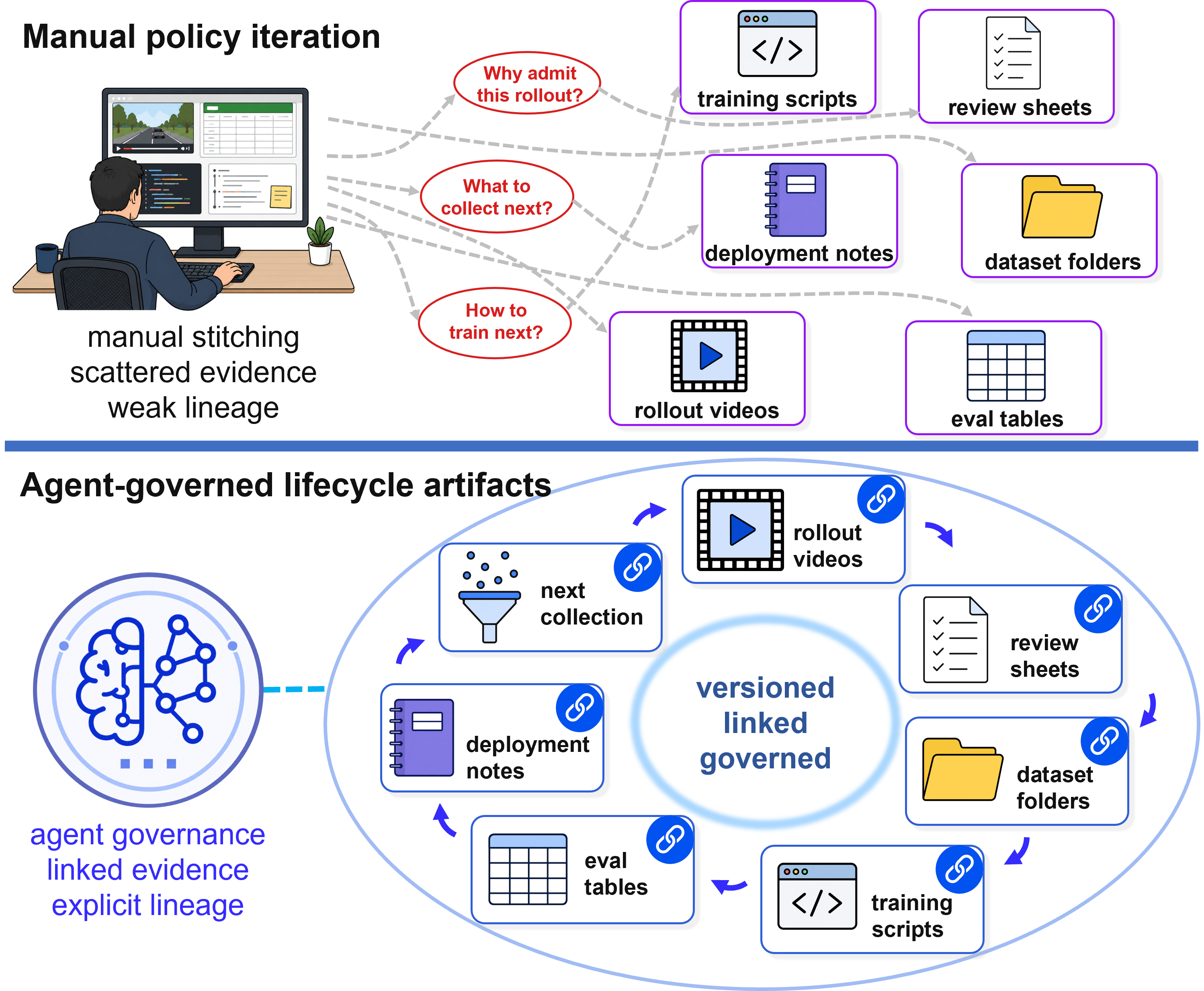}
    \caption{Manual policy iteration versus agent-governed lifecycle artifacts.}
    \label{fig:teaser}
    \vspace{-0.8em}
\end{wrapfigure}

To address this problem, we introduce \system, a lightweight, open-source,
agent-native governance system that turns the messy data ecosystem behind robot
policy iteration into an explicit data lifecycle. Each decision carries a
versioned record of its supporting evidence, so later iterations can recover
how a policy was produced. Figure~\ref{fig:teaser} summarizes this shift.
Agents write typed artifacts when they interpret a rollout or prepare data for
training. Review records link admitted episodes to evidence; training records
connect frozen datasets to checkpoints. Dataset-health reports identify gaps
before training. Evaluation summaries then carry failures into a brief for the
next collection round.

\system keeps policy iterations legible across changing robots, collection
workflows, and training stacks by preserving the same lifecycle semantics. Its
core abstraction is a lifecycle artifact contract for embodied policy
iteration: agents may interpret evidence, adapt data, and summarize state, but
consequential transitions enter the lifecycle only as validated,
evidence-linked artifacts. In routine iteration, a data-collection operator can
collect rollouts, inspect prepared evidence,
resolve uncertain cases, and launch the next training cycle through the
frontend, while \system carries the decision trail forward across policy
versions.

Our contributions are:
\begin{itemize}[leftmargin=*]
    \item We formulate robot policy iteration as a data lifecycle governance
    problem and define a typed lineage graph for tracking how evidence,
    artifacts, and decisions carry across policy iterations.
    \item We introduce an agent-native governance architecture in which agents
    operate as artifact-producing lifecycle workers, turning task-grounded
    semantic and integration judgments into bounded, auditable lifecycle state.
    \item We realize \system as a lightweight open-source implementation with
    an operator frontend and portable lifecycle artifacts, and evaluate review
    reliability, routine-cycle effort, and closed-loop recollection across
    multiple robot and training families.
\end{itemize}


\section{Related Work}
\label{sec:related}

\paragraph{Robot data collection and policy iteration.}
Large datasets give policies access to broader robot experience, while scalable
environments make it easier to expand training coverage
~\citep{dasari2019robonet,fang2023rh20t,oneill2023openx,walke2023bridgedata,khazatsky2024droid,nasiriany2024robocasa}.
Collection systems such as GELLO and UMI simplify demonstration collection and
correction
~\citep{mandlekar2018roboturk,wu2023gello,chi2024umi,chen2024arcap,fang2026robopocket}.
RoboMimic and MimicGen study how demonstrations can be used and generated for
learning~\citep{mandlekar2021robomimic,mandlekar2023mimicgen}.
Other work examines which demonstrations to retain and how dataset quality
affects policy performance
~\citep{belkhale2023dataquality,agia2025cupid,dass2026datamil,hejna2025robot,zhang2026scizor,lin2024datascaling}.
\system follows the decisions that connect these datasets to later policy
versions. It records why an episode was admitted and preserves useful failure
evidence. That history also links a policy to its training data and informs
recollection.

\paragraph{Agentic robotics systems.}
Language-model systems support affordance-grounded instruction following and
planning from observations~\citep{ahn2022saycan,huang2022inner}.
Some express robot behavior as code or 3D value maps
~\citep{liang2023code,huang2023voxposer}.
Other work develops embodied multimodal models and generalist robot policies
~\citep{driess2023palme,brohan2023rt2,kim2024openvla,octo2024,black2024pi0,black2025pi05,liu2025rdt}.
VLA fine-tuning adapts learned policies~\citep{kim2025openvlaoft}, and recent
agentic frameworks address longer planning and execution horizons
~\citep{duan2025aha,li2026roboclaw,liang2026tether}.
These systems place model agency close to robot behavior. \system applies
agentic reasoning to the data lifecycle that produces later policies. An agent
may interpret execution evidence or a training repository, but its judgment
enters the lifecycle as a schema-validated artifact.

\paragraph{MLOps, provenance, and reporting.}
ML systems research shows how technical debt and data cascades affect the
systems around a model~\citep{sculley2015hidden,sambasivan2021datacascades}.
Production pipelines and experiment trackers organize training runs and their
outputs~\citep{baylor2017tfx,zaharia2018mlflow,vartak2016modeldb,schlegel2022artifacts}.
Provenance records make data histories traceable
~\citep{souza2019provml,schlegel2025provenance}, while dataset documentation and
model cards describe assumptions and intended use
~\citep{gebru2021datasheets,mitchell2019modelcards}.
Robot policy iteration needs a comparable record of decisions grounded in
execution evidence. \system links a rollout's terminal evidence to its admission
decision and preserves subsequent overrides. Dataset locks carry those decisions
into training, and evaluation failures inform the next collection brief.


\section{Method}
\label{sec:method}

\begin{figure}[t]
    \centering
    \includegraphics[width=\linewidth]{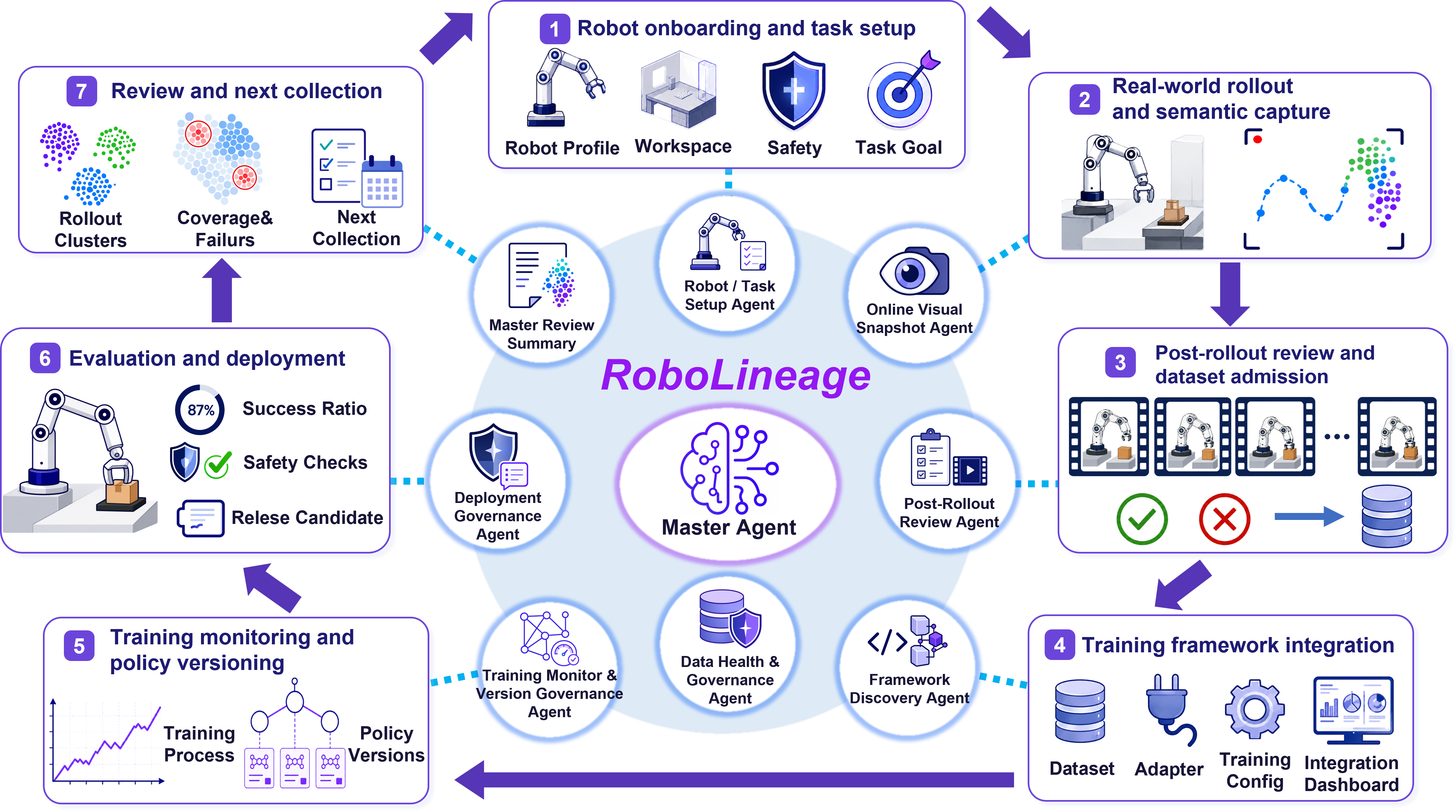}
    \caption{\system governs robot policy iteration as an explicit data
    lifecycle. The outer loop shows the policy-iteration cycle from robot/task
    grounding to next collection; the inner layer shows lifecycle agents that
    read and write typed artifacts at each transition.}
    \label{fig:overview}
\end{figure}

\subsection{Problem Definition: Governing a Robot Data Lifecycle}
\label{sec:problem}

Consider a robot policy iteration loop indexed by $k$. Let $c_k$ denote the
collection process for that iteration: a data-collection operator, teleoperation
interface, scripted routine, policy-assisted rollout protocol, intervention
policy, or a mixture of these. Under embodiment $b$ and task contract $\tau$,
$c_k$ produces rollouts $\mathcal{R}_k=\{r_i\}$. Each rollout contains raw
streams such as camera frames, robot state, actions, end-effector state,
task-relevant tool signals, synchronization metadata, and operator events. Review
turns a rollout into semantic labels and decisions: whether the task succeeded,
where it failed, whether the episode is useful for training, and whether a human
should inspect it. Accepted rollouts become dataset $D_k$, while routed and
rejected artifacts remain linked for audit, failure analysis, and later
collection. A training run $T_k$ produces policy candidate $\pi_k$; evaluation
trials $\mathcal{E}_k$ lead to deployment review, rollback, or recollection
decisions $G_k$.

We define the robot data lifecycle as the ordered set of transformations and
decisions triggered by routine operator-visible actions: demonstrate, rollout,
intervene, annotate, train, evaluate, prepare release, and revert. Abstractly,
one iteration is
\[
    (b,\tau,c_k) \rightarrow \mathcal{R}_k \rightarrow
    \mathcal{A}_k \rightarrow D_k \rightarrow T_k \rightarrow
    \pi_k \rightarrow \mathcal{E}_k \rightarrow G_k,
\]
where $\mathcal{A}_k$ denotes review and admission artifacts. Lineage records the
evidence, rules, and artifacts behind each transition. The transition is the
natural unit of governance because policy quality is often determined between
runs: which evidence was trusted, which rollout changed the dataset, which
training input produced a checkpoint, and which failure changed the next
collection plan. The typed lineage graph is the executable form of this
accountability chain: nodes carry lifecycle artifacts, and edges record the
evidence-linked transitions that allow later agents, operators, and audits to
recover why a policy version exists. If collection uses a prior policy, that
policy is recorded as part of $c_k$ and the rollout metadata, alongside other
collection sources.

We represent lineage as a typed, versioned artifact graph
\[
    \mathcal{L}_k=(\mathcal{V}_k,\mathcal{P}_k)
\]
whose nodes are lifecycle artifacts and whose edges encode relations such as
\artifact{observes}, \artifact{reviews}, \artifact{admits},
\artifact{freezes}, \artifact{trains}, \artifact{evaluates}, and
\artifact{recommends}. Setup nodes record the robot profile, task contract, and
collection context. Rollout records link raw evidence to snapshots and review
annotations; admission decisions and dataset locks specify the training inputs.
Training and evaluation nodes then track the resulting policy, with deployment
tickets and master summaries carrying decisions into the next iteration. Agents
share this graph as the governance state for successive policy versions. The outer
loop in Figure~\ref{fig:overview} visualizes this formulation from robot/task
grounding through next collection.
Appendix~\ref{app:lineage_semantics} gives the edge semantics and an end-to-end
trace.

\subsection{Agent-Native Governance Over Lifecycle Artifacts}
\label{sec:overview}

\system provides the agent-native implementation of the inner layer of
Figure~\ref{fig:overview}: agents act as artifact-producing lifecycle workers,
reading the current lineage context and writing governed artifacts at each
transition. The governance
problem is to automate enough routine work for a data-collection operator to
carry an iteration forward with agent support, while keeping data and policy
changes reproducible. This graph view motivates five design principles.
\textbf{Traceability} requires each policy to retain the evidence and data rules
that produced it. \textbf{Semantic coverage} preserves the task meaning behind
raw observations. \textbf{Agent autonomy with explicit state} lets agents make
context-dependent judgments through typed artifacts that later stages can
inspect. \textbf{Portability} localizes robot differences in profiles and learner
differences in adapters. \textbf{Operator efficiency} reduces routine review and
data preparation while retaining human inspection of ambiguous cases.

Each agent has a lifecycle scope, reads the current artifact context, writes a
governed artifact, and exposes uncertainty and dependencies to the next stage:
\[
    v_j = A_j\!\left(\mathrm{ctx}_j(\mathcal{L}_k), x_j; b,\tau\right),
    \qquad v_j \in \mathcal{T}_j ,
\]
where $v_j$ may be a visual snapshot, review annotation, data-health report,
framework profile, evaluation summary, or master summary. Agents can therefore
perform substantial semantic work while their outputs remain versioned lifecycle
state, turning context-dependent expert judgments into repeatable operations
with explicit inputs, uncertainty, and provenance.

Fixed rules implement schema checks, content hashes, dataset locks, and
configured thresholds. Agents handle context-dependent judgments: interpreting
terminal evidence, assessing whether a failure is useful, mapping a repository's
data contract, and explaining a coverage gap. Their outputs enter the lifecycle
through the same typed artifact boundary, so downstream workers can inspect the
evidence and uncertainty behind a recommendation. This design makes the
semantic work reusable across policy iterations while preserving deterministic
validation at each transition.

Appendix~\ref{app:agent_contract_pattern} describes the shared contract, and
Appendix~\ref{app:rationale} expands the module rationale.

\subsection{Robot Onboarding and Task Grounding}
\label{sec:onboarding}

During setup, \system writes a robot profile $b$ and a task contract $\tau$ into lineage.
The Robot Onboarding Agent resolves local topics and interfaces into normalized
streams. The profile records how camera, robot state, and action data are bound,
including coordinate frames and timing. Recorder health and auxiliary tool
signals support validation, and warnings remain attached to the profile. For
ROS-based systems, the operator confirms these bindings before collection.
The Task Config Agent gives the execution its task meaning. It defines phases
and success criteria, then specifies which events require closer inspection and
how episodes should be admitted. The contract also specifies the collection
policy. Downstream agents use these contracts without interpreting each robot's
local naming conventions again. Details are in
Appendix~\ref{app:robot_onboarding} and Appendix~\ref{app:task_config}.

The operator enters the task goal and recording topics through the UI. Agents
inspect topic types and sample messages, complete and validate stream bindings,
and serialize reusable profile YAML and task-contract artifacts. Goal and
success criteria seed the phase vocabulary for operator validation. A profile
therefore captures a robot's stream interface, while the task contract captures
the meanings used to interpret its execution; both can be reused across routine
cycles. This boundary keeps robot-specific naming and timing separate from
semantic review and learner-specific data preparation.

\subsection{Rollout Interpretation with Online VSA and Post-Rollout Review}
\label{sec:vsa}

Rollout interpretation is the transition where raw collection becomes lifecycle
evidence. During collection, the operator benefits from timely feedback about
phase, risk, and obvious failure. Before dataset admission, the system needs a
more careful account of outcome, terminal state, failure phase, and training
usefulness. \system therefore separates Online
Visual Snapshot Agent (VSA) from asynchronous
post-rollout review while preserving raw capture as the source of truth.

\begin{wrapfigure}{R}{0.70\linewidth}
    \vspace{-0.8em}
    \centering
    \includegraphics[width=\linewidth]{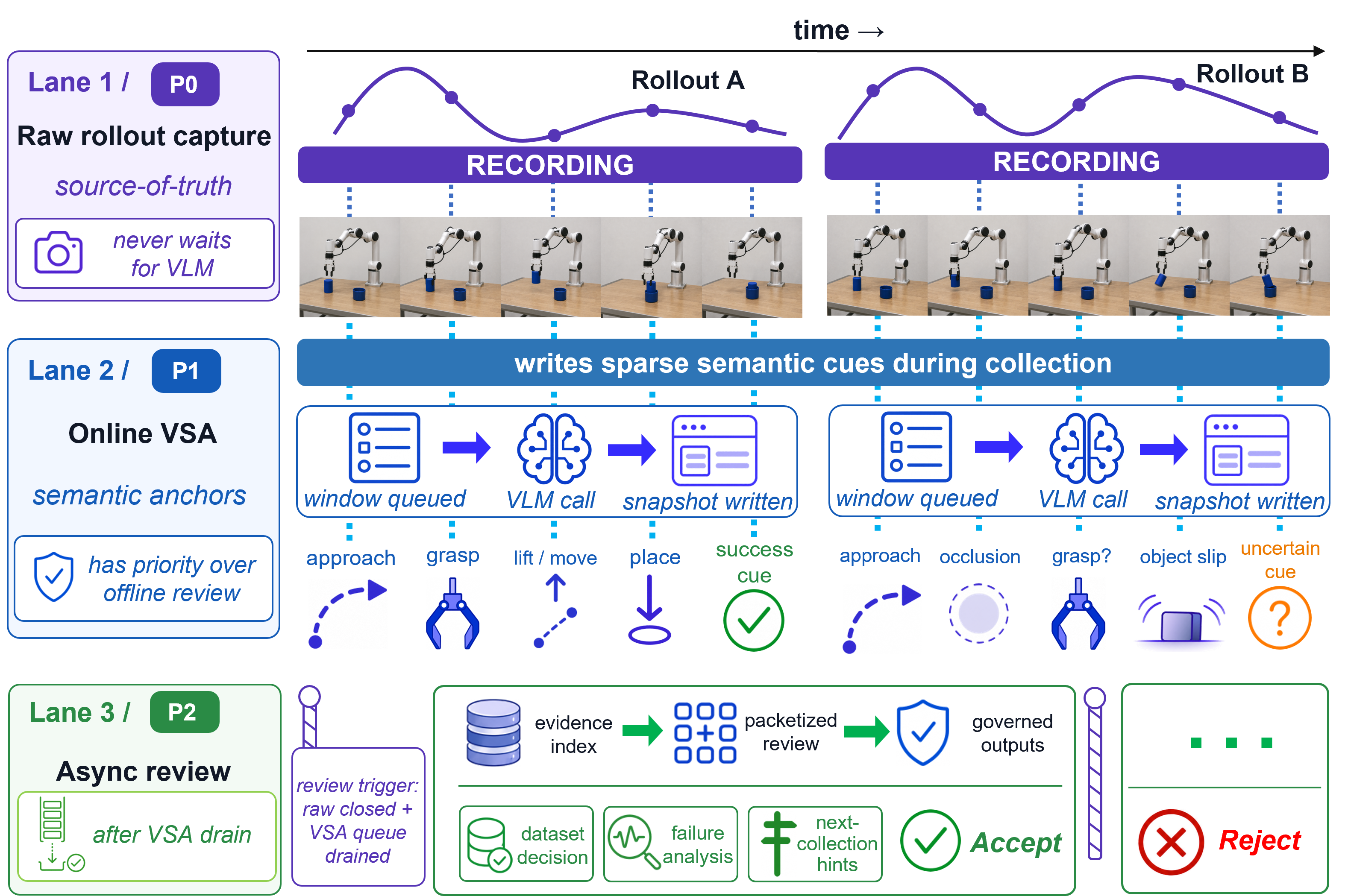}
    \caption{Rollout interpretation uses three lanes. Raw capture preserves the
    training source of truth, online VSA provides sparse semantic anchors
    during collection, and asynchronous post-rollout review turns evidence into
    governed lifecycle artifacts.}
    \label{fig:vsa_review}
    \vspace{-0.8em}
\end{wrapfigure}

The mechanism uses three coordinated lanes (Figure~\ref{fig:vsa_review}). The
raw capture lane records streams, operator events, and timing continuously, so
model timeouts or review backlogs cannot block the source of truth. The Online
Visual Snapshot Agent schedules sparse task-aware windows during collection and
writes semantic anchors for feedback, later review, and uncertainty routing.
After capture
closes and queued VSA windows drain, the
Post-Rollout Review Agent builds an evidence index and packetizes terminal
state, post-terminal stability, temporal context, and operator/log context into
a stricter final interpretation.
Appendix~\ref{app:vsa} and Appendix~\ref{app:review} give the windowing and
packet-review details.

The review lane produces a final annotation, phase timeline, failure analysis,
and dataset decision. The Data Governance Agent separates training eligibility,
human-review priority, failure-pool routing, and exclusion reasons. Only
rollouts marked \artifact{accepted\_for\_training} enter the primary training
manifest. Useful failures and ambiguous successes remain linked as governed
lineage evidence. Appendix~\ref{app:data_governance} describes the admission
fields and override trace.

VSA supplies evidence anchors and has no dataset-admission authority.
Schema-valid post-review with sufficient confidence can supersede an online
phase label while retaining both records. Severe phase drift combined with
low-confidence review triggers human inspection and withholds primary-training
admission. Training eligibility and phase-label confidence are recorded
separately, so a phase disagreement alone does not determine admission.
A model may confuse similar objects or lose evidence under occlusion. A
mismatched task contract or missing terminal observation can also lead to an
incorrect review. Schema checks catch malformed responses, and failed requests
are retried. Unresolved evidence failures receive failure or review records and
remain outside primary training. Raw
collection continues during asynchronous review.

\subsection{Data Health, Training Integration, and Version Governance}
\label{sec:training}

The next lifecycle transition turns reviewed rollouts into a training dataset
and then into a policy candidate. The Data Governance Agent first determines
whether each rollout is eligible for training. Review priority and failure-pool
routing remain separate from that decision. The Data Health Agent examines the
resulting collection, using rejected and routed episodes as well as accepted
data to identify weak task phases. Phase coverage and failure distributions show
where data are sparse. Imbalance or missing evidence can leave a dataset unready
even when individual examples were correctly admitted. The readiness artifact
records these gaps and guides later recollection.

Training integration adapts this governed dataset to the host learner. The
Framework Discovery Agent writes a framework profile from the target repository
and the user's normal commands, and the Dataset Adapter Agent materializes
accepted rollouts into that contract while preserving rollout identifiers,
review artifacts, and hashes. Adapter outputs are validated before training.
The Version Governance Agent freezes the selected episodes and adapter context
in a dataset lock before training begins. The Training Monitor Agent records
progress and metrics alongside checkpoint candidates, including warnings and
failures. Policy metadata binds the chosen checkpoint to this lock and its
parent policy; the framework profile and code revision identify how it was
trained. The adapter checks camera, state, and action streams for modality and
timestamp alignment without changing the learner's training command. These
reusable artifacts make routine preparation distinct from the
one-time work of integrating a new robot or training repository.
Details are in
Appendix~\ref{app:data_governance}, Appendix~\ref{app:dataset_health},
Appendix~\ref{app:framework_adapter}, Appendix~\ref{app:training_monitor}, and
Appendix~\ref{app:training}.

\subsection{Evaluation, Deployment Governance, and Master Synthesis}
\label{sec:governance}

\system treats evaluation and deployment as agent-interpreted lifecycle stages.
The Policy Evaluation Agent writes evaluation summaries from trial evidence:
success, failure phase, safety issue, weak task phase, and regression against
policy ancestry. The Deployment Governance Agent then converts the summary into
the next lifecycle transition: deploy, hold, roll back, or collect more.

The Master Agent provides cross-iteration synthesis. It reads compact lineage
artifacts in place of raw videos or full logs, then writes the current lifecycle
state, a human-readable review, persistent memory, and a suggested next action.
The Data Governance Agent handles rollout admission and the Data Health Agent
handles data readiness; the Master Agent turns local lifecycle artifacts into
cross-iteration memory and operator guidance. The cross-iteration node in
Figure~\ref{fig:overview} carries this context forward, explaining what changed
and which collection target the evidence supports.
See Appendix~\ref{app:eval_deploy_agents} and Appendix~\ref{app:master_review}.

\subsection{Lightweight Implementation and Frontend}
\label{sec:implementation}

\system is implemented as a lightweight backend and frontend around existing
robotics and learning stacks. The backend stores ordinary schema-validated
artifacts with content hashes. The frontend lets operators inspect rollouts and
resolve review decisions. It also exposes dataset readiness and follows training
through evaluation to deployment review. The Master Agent view presents the next
collection brief alongside the current iteration state. The release includes
the runtime and schemas, with sample artifacts and scoring scripts for
reproduction. Representative prompts document the agents' contracts.
ROS2-based robots can be adapted through profile-guided
bindings for camera, state, action, and task-relevant streams.
Appendix~\ref{app:frontend} describes the operator-facing flow;
Appendix~\ref{app:implementation} lists the ordered components;
Appendix~\ref{app:release} describes the release package and reproduction
boundaries, and summarizes the core prompt contracts.


\section{Experiments}
\label{sec:experiments}

We evaluate \system around existing policy learners with three questions: (i)
does the VSA/post-review pipeline improve rollout review and admission, (ii)
does the lifecycle interface preserve policy quality while reducing routine
pre-training labor, and (iii) can lineage-guided recommendations drive targeted
recollection?

\subsection{Experimental Setup}

\begin{wraptable}{R}{0.60\linewidth}
    \vspace{-1.05\baselineskip}
    \centering
    \scriptsize
    \setlength{\abovecaptionskip}{1pt}
    \setlength{\belowcaptionskip}{2pt}
    \setlength{\tabcolsep}{2.2pt}
    \caption{VSA and post-rollout review over 25 tasks and 500 rollouts.
    All values are percentages.
    Outcome and failure-phase accuracy measure semantic review. Admission
    accuracy is exact agreement with expert consensus over normalized data-use
    routes across all 500 rollouts, before operator overrides. Leakage is the
    percentage of 194 expert-rejected rollouts admitted
    to the primary training manifest.
    Bold marks the best automated method. $^\dagger$Human agreement is a
    pre-consensus inter-annotator reference on the same 500-rollout benchmark,
    so leakage is not applicable.}
    \label{tab:semantic}
    \begin{tabular}{p{0.31\linewidth}cccc}
        \toprule
        Method & \shortstack{Outcome\\acc. $\uparrow$} &
        \shortstack{Failure-phase\\acc. $\uparrow$} &
        \shortstack{Admission\\acc. $\uparrow$} & Leakage $\downarrow$ \\
        \midrule
        Direct video-to-VLM & 78.4 & 58.0 & 79.6 & 16.0 \\
        VLM + schema only & 82.6 & 63.4 & 84.0 & 11.3 \\
        VSA-only (no post-review) & 84.0 & 68.0 & 85.6 & 9.8 \\
        Post-review only (no VSA) & 86.6 & 72.0 & 88.4 & 7.2 \\
        No final observation / terminal focus & 89.0 & 78.0 & 91.0 & 6.2 \\
        No packetized review & 92.0 & 83.6 & 93.4 & 4.6 \\
        Full \system & \textbf{96.0} & \textbf{88.0} & \textbf{96.4} & \textbf{2.1} \\
        \midrule
        \emph{Human agreement reference}$^\dagger$ & \emph{97.0} & \emph{91.0} & \emph{98.0} & -- \\
        \bottomrule
    \end{tabular}
    \vspace{-0.35\baselineskip}
\end{wraptable}

Experiments cover ARX with ACT-style action chunking, Realman with Diffusion
Policy, and GALBOT G1 with a VLA/LeRobot-style workflow
(Appendix Figure~\ref{fig:robots_app}). The tasks are tabletop or
visually grounded manipulation settings where cameras and robot state provide
the main evidence. The review benchmark contains 25 tasks and 500 rollouts. The
closed-loop case studies use three recollection rounds and an independent
60-trial final test. Labeling, task-contract, timing, and model-backend
protocols are in
Appendix~\ref{app:protocol}.

\subsection{Review Reliability and Ablations}
\label{sec:semantic_results}

Table~\ref{tab:semantic} isolates the review pipeline. Baselines use the same
rollout pool and model route but remove the artifact structure, online anchors,
packetization, or terminal evidence; Appendix~\ref{app:protocol} defines the
annotation rubric and baseline controls, and Appendix~\ref{app:review_benchmark}
summarizes the 25-task coverage.

Full \system reaches the best automated outcome and admission accuracy, the
best automated failure-phase localization, and the lowest bad-data leakage. Direct
video review is weakest on failure phase and false admission;
schema-only review improves formatting but lacks event anchors; VSA-only review
is useful for feedback and indexing but not strict enough for final admission.
The post-review-only and evidence-ablation rows measure the semantic-evidence
path: online anchors, packetized review, and terminal evidence improve review
under the same model and rollout pool. Cross-round memory is evaluated separately
in Section~\ref{sec:policy_results}. Before operator overrides, the full pipeline
matches consensus on 482 of 500 routes; 74 cases are routed for inspection and
18 later receive operator changes.
Appendix~\ref{app:protocol} details baseline definitions, routing
diagnostics, and containment cases
(Tables~\ref{tab:baseline_definitions}, \ref{tab:routing_diagnostics},
and~\ref{tab:failure_examples}). Appendix~\ref{app:review_benchmark} gives the
task roster, per-family and per-robot breakdowns, model-backend replay, and
qualitative VSA/post-review plates
(Tables~\ref{tab:task_roster}, \ref{tab:per_family_semantic},
\ref{tab:per_robot_semantic}, and~\ref{tab:model_sensitivity};
Figures~\ref{fig:plate_arx}--\ref{fig:plate_galbot}).

\subsection{Policy Quality and Effort}
\label{sec:effort_results}

\begin{figure}[t]
\begin{minipage}[t]{0.48\linewidth}
\vspace{0pt}

    \centering
    \setlength{\abovecaptionskip}{2pt}
    \setlength{\belowcaptionskip}{1pt}
    \begin{tikzpicture}
    \begin{axis}[
        ybar,
        width=\linewidth,
        height=0.17\textheight,
        ymin=0,
        ymax=105,
        ylabel={Succ. (\%)},
        ylabel style={font=\scriptsize,xshift=1.1em},
        yticklabel style={font=\tiny,xshift=0.3em},
        symbolic x coords={Block-swap,Pick-reset,Cup,Sorting,Bottle,Tissue,Tool,Mobile},
        xtick=data,
        xticklabels={Block,Pick,Cup,Sort,Bottle,Tissue,Tool,Mobile},
        x tick label style={rotate=45,anchor=east,font=\tiny},
        tick label style={font=\tiny},
        label style={font=\scriptsize},
        bar width=2.2pt,
        enlarge x limits=0.10,
        clip=false,
        legend style={font=\tiny,at={(0.5,1.10)},anchor=south,legend columns=2,draw=none},
        ymajorgrids=true,
        grid style={gray!20},
    ]
    \addplot+[error bars/.cd, y dir=both, y explicit]
    coordinates {(Block-swap,96.7) +- (0,2.3) (Pick-reset,95.0) +- (0,2.8) (Cup,93.3) +- (0,3.2) (Sorting,96.7) +- (0,2.3) (Bottle,91.7) +- (0,3.6) (Tissue,90.0) +- (0,3.9) (Tool,93.3) +- (0,3.2) (Mobile,88.3) +- (0,4.1)};
    \addplot+[error bars/.cd, y dir=both, y explicit]
    coordinates {(Block-swap,95.0) +- (0,2.8) (Pick-reset,93.3) +- (0,3.2) (Cup,93.3) +- (0,3.2) (Sorting,95.0) +- (0,2.8) (Bottle,91.7) +- (0,3.6) (Tissue,91.7) +- (0,3.6) (Tool,93.3) +- (0,3.2) (Mobile,88.3) +- (0,4.1)};
    \legend{Human expert workflow,RoboLineage workflow}
    \end{axis}
    \end{tikzpicture}

    \caption{\textbf{Policy quality.} Each task uses a shared 100-rollout candidate pool.
    Each workflow trains on 60 selected demonstrations and evaluates on 60 trials per
    task; error bars show binomial standard error. Success counts agree on four
    tasks and differ by one trial on the other four. Mean success is 93.1\% for
    the expert workflow and 92.7\% for \system.}
    \label{fig:quality_bar}

\end{minipage}
\hfill
\begin{minipage}[t]{0.48\linewidth}
\vspace{0pt}

    \centering
    \begin{tikzpicture}
    \begin{axis}[
        ybar stacked,
        width=\linewidth,
        height=0.15\textheight,
        ymin=0,
        ymax=145,
        ylabel={Minutes},
        symbolic x coords={Human,RoboLineage},
        xtick=data,
        tick label style={font=\tiny},
        label style={font=\scriptsize},
        bar width=12pt,
        clip=false,
        legend style={font=\tiny,at={(0.50,0.58)},anchor=center,
        legend columns=2,draw=none,fill=white,fill opacity=0.75,
        text opacity=1},
        ymajorgrids=true,
        grid style={gray!20},
    ]
    \addplot coordinates {(Human,10.1) (RoboLineage,2.3)};
    \addplot coordinates {(Human,40.5) (RoboLineage,40.6)};
    \addplot coordinates {(Human,59.2) (RoboLineage,9.4)};
    \addplot coordinates {(Human,20.4) (RoboLineage,2.0)};
    \legend{Onboard,Collection,Review,Integration}
    \end{axis}
    \end{tikzpicture}

    \caption{\textbf{Routine-cycle effort.} Mean active pre-training time per 30-rollout cycle over 12
    matched sessions. \system reduces review and integration labor.}
    \label{fig:effort_bar}

    \captionof{table}{\textbf{Runtime and API footprint.} Costs are estimated from logged
    token/image usage for one 30-rollout iteration.}
    \label{tab:runtime_cost}

    \tiny
    \resizebox{\linewidth}{!}{%
    \begin{tabular}{lcccc}
        \toprule
        Stage & When & Median & p90 & Cost/cycle \\
        \midrule
        VSA snapshot & During rollout & 2.1 s & 4.8 s & \$6.10 \\
        Post-review & Async & 18.4 s & 34.0 s & \$4.80 \\
        Summaries & After session & 7.2 s & 13.6 s & \$0.70 \\
        \midrule
        30-rollout cycle & Mixed & -- & -- & \$11.60 \\
        \bottomrule
    \end{tabular}
    }

\end{minipage}
\end{figure}

We ask whether \system can reduce expert-mediated lifecycle labor while matching
the policy quality of an expert-managed workflow in the measured setting. For
each of eight tasks distinct from the closed-loop studies, we record a fixed
100-rollout candidate pool while \system's online VSA runs during collection.
The expert-managed workflow and \system then independently select 60 accepted
demonstrations from the same pool, using the same policy family, training
command, and evaluation protocol.
Mean success is 92.7\% for \system and 93.1\% for the expert-managed workflow,
a mean difference of 0.4 percentage points
(Figure~\ref{fig:quality_bar}). The workflows have equal success counts on four
tasks; the remaining four differ by one trial out of 60. Three of these favor
the expert workflow and one favors \system. These results show comparable
policy performance when both workflows prepare training sets from the same
candidate pool. Appendix~\ref{app:policy_quality_protocol} gives protocol details.

Two trained operators run both workflows over 12 matched 30-rollout sessions,
four per robot family, with condition order alternated after practice.
The manual baseline uses the lab's maintained scripts, review sheets, and
conversion workflow. Review includes full-video viewing, phase and transition
marking, outcome/failure diagnosis, rationale, and training-manifest assembly:
59.2 minutes per session, or 1.97 minutes per rollout. Online VSA runs during
recording, and asynchronous post-review overlaps subsequent collection;
The 9.4-minute review stage in \system measures residual blocking work, uncertain-case
inspection, and manifest confirmation. Active time also includes onboarding and
framework preparation, with physical collection similar at 40.5 and 40.6 minutes.
Non-collection effort falls from 89.7 to 13.7 minutes; total effort falls from
130.2 to 54.3 minutes, a 2.4$\times$ reduction. These routine timings follow
one-time robot/learner integration and exclude GPU training time.
Appendix~\ref{app:effort_protocol} and Tables~\ref{tab:effort_scope}
and~\ref{tab:effort_variance} give scope and variance.
Table~\ref{tab:runtime_cost} reports \$11.60 in hosted model calls per cycle.

\subsection{Closed-Loop Recollection Case Studies}
\label{sec:policy_results}

Lineage-guided improvement turns failure analysis into data-collection action.
We test this mechanism on two closed-loop real-robot tasks: cube stacking probes
visual coverage gaps, and drawer opening probes contact-rich phase transitions.
Each method is evaluated over three independent runs. In each run, three
recollection rounds each add 30 accepted demonstrations and use 30 diagnostic
trials. After the third round, the policy receives an independent 60-trial final
test. The initial policy $\pi_0$ is also evaluated on
60 trials per run, for 180 trials per task. The heuristic
counts recent failure phases; \system compares current failures with accepted
data and cross-round lifecycle state to decide what to collect next.
\system tracks human-expert guidance and outperforms the other planners
(Table~\ref{tab:policy}). All planners receive the current diagnostic rollouts. The no-memory
planner uses the same current VSA/post-review evidence. Its planning context
is reset between rounds to exclude earlier lifecycle memory and policy ancestry.

\begin{wraptable}{R}{0.62\linewidth}
    \vspace{-0.8\baselineskip}
    \centering
    \scriptsize
    \setlength{\tabcolsep}{2.2pt}
    \caption{Closed-loop recollection trajectories. Each cell reports mean
    success rate (\%) over three independent runs. Initial and final policies use
    60 trials per run; Iter.1--3 use 30. Iter.1/Iter.2 appear on the first line
    and Iter.3/Final on the second. Phase denotes the failure-phase heuristic;
    No memory denotes the no-lineage-memory planner. Bold marks \system.}
    \label{tab:policy}
    \resizebox{\linewidth}{!}{%
    \begin{tabular}{@{}lccccc@{}}
        \toprule
        Task & Random & Phase & No memory & \system & Human expert \\
        \midrule
        \shortstack[l]{Stack\\($\pi_0$=25.0)} &
        \shortstack{30.0/33.3\\37.8/36.7} &
        \shortstack{44.4/56.7\\67.8/66.7} &
        \shortstack{47.8/60.0\\72.2/70.0} &
        \shortstack{\textbf{52.2/68.9}\\\textbf{84.4/83.3}} &
        \shortstack{55.6/72.2\\87.8/86.7} \\
        \shortstack[l]{Drawer\\($\pi_0$=26.7)} &
        \shortstack{31.1/34.4\\37.8/37.8} &
        \shortstack{42.2/53.3\\63.3/62.2} &
        \shortstack{45.6/57.8\\67.8/66.7} &
        \shortstack{\textbf{50.0/65.6}\\\textbf{80.0/78.9}} &
        \shortstack{53.3/68.9\\83.3/81.1} \\
        \bottomrule
    \end{tabular}
    }
    \vspace{-0.6\baselineskip}
\end{wraptable}

The recommendations target the dominant failure mode at each round. In stacking,
they move from out-of-distribution grasp
locations to centered pre-release placement and then bounded-dwell release
examples. Relative to the no-memory planner, final success increases from
70.0\% to 83.3\% in stacking and from 66.7\% to 78.9\% in drawer opening.
Appendix~\ref{app:closed_loop_protocol} gives the full recollection
protocol, and Appendix~\ref{app:closed_loop_cases} gives the visual evidence
plate and round-by-round recommendations.

\subsection{Storage and Local Model Serving}
\label{sec:storage_local}

Across the 500-rollout benchmark, raw records average 72.6 MiB per rollout.
Structured lineage adds 68 KiB (0.09\% of raw storage); the regenerable VSA
keyframe cache adds 16.3 MiB (22.5\%). We also replay the fixed 200-rollout backend subset locally
with Qwen3-VL-32B-Instruct through vLLM on two RTX 4090 GPUs modified to 48 GB
each, using the same prompts, contracts, schemas, and thresholds without
retuning. Outcome, phase, and admission accuracy are 82.5\%, 73.0\%, and 83.5\%;
false accepts are 6/100 and review routes 53/200.
Local serving provides an open-weight route with lower accuracy and more review
than the strongest hosted backend. It removes per-call API fees; amortized cost
depends on hardware utilization. Appendix~\ref{app:storage_footprint} and
Table~\ref{tab:model_sensitivity} give storage accounting and backend results.


\section{Limitations and Discussion}
\label{sec:limitations}

\system reduces routine workflow effort and achieves policy quality comparable
to expert-managed workflows under matched budgets. It also uses failure
analysis to guide recollection. Our evidence comes from three configured
robot/learner pairs on visually grounded manipulation tasks, with existing
policy learners and lab safety procedures. Transfer to navigation or tasks
with little visual evidence remains untested.

The ablations examine semantic review and memory across policy iterations.
Provenance records make decisions traceable; dataset locks and deployment
tickets govern training and release. We evaluate these mechanisms together
and have not isolated their individual contributions.

Admission decisions can still go wrong if the model misreads an observation
or the task contract is incomplete. Missing terminal evidence may also force
a rollout into human review. As workload grows, the review queue could become
a bottleneck. Sustained operation with thousands of rollouts therefore needs
latency and backlog measurements. Distributed fleets and edge deployment have
not been tested. Appendix~\ref{app:future} discusses these limitations and future
directions.

\clearpage
\section*{Acknowledgments}

This work is supported by the Early Career Scheme of the Research Grants Council (RGC) grant \# 27207224, the HKU-100 Award, a donation from the Musketeers Foundation, and in part by the JC STEM Lab of Autonomous Intelligent Systems funded by The Hong Kong Jockey Club Charities Trust and computing resource provided by HKU-Shanghai Intelligent Computing Research Center.

We thank the robot operators and collaborators who supported data collection, system testing, and real-robot experiments.

\bibliography{dblq}

\clearpage
\appendix

\section{System Implementation Details}
\label{app:implementation}

\begin{table}[h]
    \centering
    \small
    \begin{tabular}{p{0.23\linewidth} p{0.32\linewidth} p{0.34\linewidth}}
        \toprule
        Agent & Role & Main lineage artifact \\
        \midrule
        Robot Onboarding & Normalize embodiment profile, streams, frames, and capabilities & \artifact{robot.profile.yaml} \\
        Task Config & Define phases, success criteria, risks, and collection policy & \artifact{task.config.json} \\
        Online Visual Snapshot & Convert sparse execution events into stable semantic evidence & \artifact{snapshot.jsonl}, keyframes \\
        Post-Rollout Review & Aggregate multimodal evidence into outcome, phase timeline, failure analysis, and L1 annotation & \artifact{annotation.final.json} \\
        Data Governance & Route rollouts to training, review, failure pool, or rejection & \artifact{dataset\_decision.json} \\
        Data Health & Summarize coverage, phase balance, missing evidence, and failure distribution & \artifact{dataset.health.json} \\
        Framework Discovery & Discover training command, data contract, and adapter path & \artifact{framework.profile.json} \\
        Dataset Adapter & Materialize accepted rollouts for the target learner & \artifact{dataset\_manifest.json} \\
        Training Monitor & Track training progress, errors, and metrics & \artifact{train.run.json} \\
        Version Governance & Lock the dataset and bind policy metadata & \artifact{dataset.lock.json}, \artifact{policy.meta.json} \\
        Policy Evaluation & Review evaluation trials and aggregate quality & \artifact{eval.summary.json} \\
        Deployment Governance & Recommend deploy, hold, rollback, or collect more & \artifact{deploy.ticket.json} \\
        Master Agent & Maintain lifecycle state, memory, risks, and next-collection summaries & \artifact{master.state.json}, \artifact{master.memory.jsonl}, \artifact{master.review.json} \\
        \bottomrule
    \end{tabular}
    \caption{Ordered agent list. The main paper introduces the
    lifecycle contract; this appendix records the implementation-level roles and
    artifacts.}
    \label{tab:agents_app}
\end{table}

\subsection{Artifact Store}
\label{app:artifact_store}

\system stores lifecycle state as structured artifacts with explicit schemas and
hashes. This format fits robotics labs: artifacts can be versioned,
inspected, copied with datasets, and debugged with ordinary tools. Each artifact
records a schema version, creation time, parent identifiers, upstream content
hashes, and a compact human-readable summary. The most common artifacts are:
\begin{itemize}[leftmargin=*]
    \item \artifact{robot.profile.yaml}: robot embodiment, topic bindings,
    cameras, state streams, action interfaces, frames, safety limits, and
    onboarding warnings.
    \item \artifact{task.config.json}: task phases, success criteria, failure
    criteria, risk events, snapshot policy, collection policy, and admission
    rules.
    \item \artifact{snapshot.jsonl}: online VSA snapshots, including event
    window, selected keyframes, predicted phase, progress, risk flags,
    uncertainty, and evidence references.
    \item \artifact{annotation.final.json}: final rollout outcome, phase
    timeline, L1 annotation, failure analysis, evidence packets, and review
    confidence.
    \item \artifact{dataset\_decision.json}: structured admission decision,
    reason codes, review routing, training eligibility, and failure-pool
    routing.
    \item \artifact{dataset.health.json}: coverage, imbalance, missing data,
    failure distribution, and dataset-level warnings.
    \item \artifact{framework.profile.json}: discovered training entry point,
    expected data contract, command templates, adapter strategy, and repository
    notes.
    \item \artifact{dataset.lock.json}: immutable manifest of episodes,
    artifact hashes, adapter version, and task/robot hashes used by a training
    run.
    \item \artifact{policy.meta.json}: policy identifier, parent policy,
    checkpoint hash, dataset lock hash, training metrics, code revision, and
    framework profile.
    \item \artifact{eval.summary.json}: evaluation trials, success metrics,
    failure phases, safety issues, regressions, and evidence links.
    \item \artifact{deploy.ticket.json}: advisory deployment recommendation,
    threshold checks, approval state, policy hash, and rollback target.
\end{itemize}

\subsection{Agent Contract Pattern}
\label{app:agent_contract_pattern}

All agents follow a common contract: structured inputs, schema-validated
outputs, explicit uncertainty, and a declared lifecycle scope. An agent may call
a multimodal or language model, and the result enters lineage only after schema
validation. Across the system, model outputs provide semantic perception,
agents translate that perception into task-grounded interpretation, and lineage
artifacts carry the versioned governance state.

\subsection{Open-Source Release and Key Prompt Contracts}
\label{app:release}
\label{app:prompt_contracts}

The release includes the backend runtime and frontend console. Schemas define
the artifacts they exchange, and representative prompts document the agents'
contracts. Tests and sample artifacts support inspection; analysis scripts
regenerate the paper tables. Hardware profiles and training-framework examples
serve as replaceable adapters. The stable release boundary is the lifecycle
artifact contract.

The scoring scripts in the package regenerate the reported metrics from frozen
labels, frozen model outputs, sampled keyframes, and hashed raw manifests,
without requiring private full-resolution videos. Re-running semantic agents
from raw evidence requires an equivalent model route and API access, so the
package separates artifact-level reproduction from model-backend reproduction.
From released artifacts, reviewers can regenerate the aggregate review tables,
timing summaries, model-call logs, and closed-loop tables. Re-training policies,
re-running robots, or auditing every full-resolution video requires local
hardware or access to private lab data.

\begin{table}[h]
    \centering
    \scriptsize
    \begin{tabular}{p{0.24\linewidth}p{0.66\linewidth}}
        \toprule
        Package item & Contents and purpose \\
        \midrule
        Backend runtime & Session service, raw capture orchestration,
        online VSA, post-rollout review, dataset governance, training
        integration, evaluation, deployment governance, and Master Agent code. \\
        Frontend console & Operator-facing lifecycle views for onboarding,
        rollout capture, review, dataset health, training integration, and
        master summaries. \\
        Artifact schemas & JSON/YAML schemas for snapshots, annotations,
        dataset decisions, dataset locks, policy metadata, evaluation
        summaries, deployment tickets, and next-collection briefs. \\
        Prompts & Frozen representative prompt templates for VSA,
        post-rollout packet review, Data Health Agent interpretation, framework
        discovery, deployment governance, and master summaries. \\
        Sample records & A minimal lifecycle trace from rollout
        capture through review, dataset admission, dataset lock, policy
        metadata, evaluation summary, and deployment recommendation. \\
        Frontend and packet examples & Review-screen screenshots, evidence
        packet excerpts, operator override records, and the corresponding
        artifact JSON for the same rollouts. \\
        Benchmark evidence & Consensus labels, sampled keyframes, evidence
        packets, and packet-review outputs for reproducing review metrics
        without redistributing full robot videos. \\
        Tests and scripts & Unit tests for contracts and agents, plus scripts
        for semantic-review scoring, effort aggregation, and closed-loop case
        tables. \\
        \bottomrule
    \end{tabular}
\caption{Contents of the open-source \system release.}
    \label{tab:release_package}
\end{table}

\begin{table}[h]
    \centering
    \scriptsize
    \begin{tabular}{p{0.22\linewidth}p{0.26\linewidth}p{0.41\linewidth}}
        \toprule
        Example artifact & Representative content & Downstream use \\
        \midrule
        rollout-042/raw\_manifest.json & camera stream, robot-state
        stream, operator stop event, frame hashes & source of truth for review
        and training adapters \\
        rollout-042/snapshot.jsonl & five VSA windows:
        approach, grasp, lift, place, final observation & semantic anchors for
        packetized post-review \\
        \path{rollout-042/review_packet_terminal.json} & terminal frames,
        VSA phase anchors, stop reason, confidence, cited frame IDs & concrete
        evidence shown in the review screen and consumed by post-rollout review \\
        annotation.final.json & outcome = failure,
        phase = grasp, evidence = gripper-center offset & input to dataset
        admission and failure analysis \\
        dataset\_decision.json & decision =
        failure-pool candidate, accepted\_for\_training=false &
        prevents bad-data leakage while preserving failure evidence \\
        dataset.health.json & OOD grasp-location gap,
        under-covered start bins & target for next recollection \\
        next\_collection\_brief.json & collect successful
        grasps in underrepresented table regions & operator-facing action for
        the next rollout batch \\
        policy.meta.json & parent policy, dataset lock, adapter
        hash, checkpoint hash & policy ancestry for evaluation and deployment
        tickets \\
        \bottomrule
    \end{tabular}
    \caption{A lifecycle trace included in the open-source release. The
    example follows one rollout-level failure as it becomes a failure-pool
    artifact, Data Health Agent gap, next-collection action, and policy-ancestry
    record.}
    \label{tab:artifact_trace}
\end{table}

\paragraph{Online VSA prompt contract.}
The online VSA prompt binds a short image window to the task contract, current
rollout memory, and auxiliary action signals. The prompt asks the VLM to choose
one configured phase and return a compact JSON assessment.

\noindent\fbox{\begin{minipage}{0.94\linewidth}
\footnotesize
\textbf{Role.} Embodied visual assessment for robot task execution monitoring.\\
\textbf{Inputs.} Task description; allowed phases; phase definitions and visual
signatures; known failure signals; recent rollout memory; event type; anchor
frame; timestamps; selected keyframes; optional gripper, end-effector, stillness,
and motion summaries.\\
\textbf{Rules.} Use visual evidence for object state, contact, placement, and
failure risk. Use action signals to refine timing and disambiguate contact,
release, retry, or stall. Select phases only from the task contract, and keep
the sequence temporally coherent with previous snapshots.\\
\textbf{JSON fields.} \artifact{phase}, \artifact{progress},
\artifact{risk\_level}, \artifact{imminent\_failure},
\artifact{needs\_review}, and \artifact{confidence}.
\end{minipage}}

\paragraph{Post-rollout review prompt contract.}
Post-rollout review is scoped to evidence interpretation. It reviews selected
evidence packets and returns local visual support that the aggregator combines
with deterministic timelines and failure rules.

\noindent\fbox{\begin{minipage}{0.94\linewidth}
\footnotesize
\textbf{Single-prompt path.} Return final success, confidence, final phase,
terminal evidence, phase corrections, retry/failure evidence, label quality,
training usability, and concise reasoning.\\
\textbf{Packet path.} For terminal, post-terminal, or temporal packets, return
packet terminal state, terminal intactness, post-terminal status, confidence,
local final phase, evidence frames, retry/failure evidence, and reasoning.\\
\textbf{Rules.} Treat retry as an event. Leave dataset admission to the
governance stage. Advisory \artifact{training\_usability} is later
separated from \artifact{accepted\_for\_training}. If visual evidence is missing,
report insufficient evidence.
\end{minipage}}

\paragraph{Robot onboarding prompt contract.}
The onboarding prompt turns a newly attached robot profile into a compact
operator-facing explanation of the available streams, frames, capabilities, and
warnings. It does not invent bindings; it explains and checks the profile that
will become the lifecycle entry point.

\noindent\fbox{\begin{minipage}{0.94\linewidth}
\footnotesize
\textbf{Inputs.} Normalized robot profile, operator note, deterministic warnings,
and deterministic assumptions.\\
\textbf{Output.} Profile summary, binding explanation, warnings, assumptions,
and recommended validation checks.\\
\textbf{Rule.} Use only the ROS topics, streams, frames, and capabilities in the
supplied profile. Profile activation and signal validation remain runtime
operations.
\end{minipage}}

\paragraph{Data Governance Agent and Data Health Agent prompt contract.}
These prompts operate after semantic review has produced evidence. Data
Governance records rollout-level routing, while Data Health summarizes the
dataset-level gaps that matter for training readiness and recollection.

\noindent\fbox{\begin{minipage}{0.94\linewidth}
\footnotesize
\textbf{Inputs.} Selected rollouts, post-review artifacts, dataset history, task
phases, deterministic coverage report, and failure-type counts.\\
\textbf{Output.} Dataset health summary, coverage notes, risk notes,
recommended collection object, and confidence.\\
\textbf{Rule.} Explain missing phase coverage, imbalance, and targeted failure
collection needs using only supplied artifacts. The deterministic report remains
the auditable source for training readiness.
\end{minipage}}

\paragraph{Framework Discovery Agent and Training Monitor Agent prompt contracts.}
Framework discovery translates a training repository and the user's normal
commands into a lifecycle adapter plan. Training monitoring then explains logged
training state without changing the training code or policy checkpoint.

\noindent\fbox{\begin{minipage}{0.94\linewidth}
\footnotesize
\textbf{Framework Discovery Agent inputs.} Repository tree, user-provided
dataset/train/eval commands, command context, inferred target data contract,
deep-inspection signals, and selected code snippets.\\
\textbf{Framework Discovery Agent output.} Framework type, target dataset expectation,
adapter plan, training/evaluation entry points, output locations, log patterns,
assumptions, and warnings.\\
\textbf{Training Monitor Agent output.} Diagnosis, likely causes, recommended action,
operator brief, and confidence derived from deterministic log metrics and a
bounded log excerpt. The monitor explains the deterministic status artifact.
\end{minipage}}

\paragraph{Policy Evaluation Agent, Deployment Governance Agent, and Master Agent prompt contracts.}
These prompts close an iteration. Evaluation produces policy-level evidence,
Deployment Governance turns that evidence into an auditable recommendation, and
the Master Agent maintains the cross-iteration summary used for the next cycle.

\noindent\fbox{\begin{minipage}{0.94\linewidth}
\footnotesize
\textbf{Policy Evaluation Agent.} Reuse the post-rollout evidence path, then write
policy-evaluation and collection-recommendation artifacts.\\
\textbf{Deployment Governance Agent.} Read deterministic evaluation summaries,
deployment decisions, collection recommendations, and next-collection briefs;
return diagnostic risk notes, an operator brief, confidence, and an
interpretation of the lifecycle transition.\\
\textbf{Master Agent.} Read compact lifecycle artifacts and return a global
summary, operator brief, risk interpretation, suggested next action, and memory
updates. The output is reviewable lifecycle state.
\end{minipage}}

\subsection{Collection Workflow Boundary}
\label{app:collection_boundary}

The evaluated implementation focuses on governed rollout capture. Raw streams
are recorded as source evidence, event-triggered VSA snapshots provide online
semantic anchors, and the Post-Rollout Review Agent writes the authoritative
annotation.
The task config also records collection policy, so future correction and
deployment-time collection modes can reuse the same lineage contract. We leave
those richer modes to future work.

\section{Detailed Agent Descriptions}
\label{app:agents}

\subsection{Robot Onboarding Agent}
\label{app:robot_onboarding}

\paragraph{Role.}
The Robot Onboarding Agent normalizes robot-specific facts into a common profile. It is
the robot-side analogue of the Framework Discovery Agent: local interface assumptions
become a portable lifecycle contract.

\paragraph{Inputs and output.}
The agent reads a YAML profile draft serialized from the operator's UI task/topic
inputs, together with deployment notes, calibration metadata, and stream
descriptions. It completes and validates bindings and emits a reusable robot
profile, validation report, warnings, and optional binding explanation
used by VSA, post-review, dataset conversion, evaluation, deployment governance,
and the frontend.

\paragraph{Profile-guided binding procedure.}
The onboarding pass normalizes the supplied profile, checks required fields, and
validates semantic bindings for camera, state, action, gripper, contact, health,
and optional operator-event channels. For ROS-based systems, the profile records
topic names, message types, transport, namespace, domain id, and field mappings
for runtime binding. The agent explains binding risks and recommends checks; the
runtime remains profile-guided.

\paragraph{Operator correction.}
The agent records assumptions and warnings, exposes suspicious or incomplete
bindings to the frontend, and waits for profile activation and signal validation
before downstream collection. Corrected bindings are written back into the robot
profile, so later rollouts are interpreted against the contract actually used in
the session.

\paragraph{Profile contents.}
The profile records embodiment type, controllable joints or end-effector
interfaces, camera streams, state topics, action topics, gripper or contact
signals, coordinate frames, safety limits, stream synchronization status, and
warnings.

\paragraph{Lifecycle behavior.}
A data-collection operator can approve or correct bindings in the UI.
Downstream modules consume normalized fields, with robot-specific topic names
and file layouts resolved by the profile. Topic bindings are stored as data in
the onboarding artifact: when a robot exposes a different camera name or state
channel, the artifact changes and the rest of the lifecycle keeps the same
interface.

\paragraph{Failure traceability.}
Onboarding is part of lineage. If a later policy fails because a stream was
missing, mis-bound, or low confidence, the team can trace the failure to the
profile used by that lifecycle run. Multi-robot portability comes from making
robot differences explicit and versioned.

\subsection{Task Config Agent}
\label{app:task_config}

\paragraph{Role.}
The Task Config Agent converts task intent into operational lifecycle rules. It defines
success, risk, failure, and training usefulness for a specific policy iteration.

\paragraph{Contract contents.}
The task artifact records phases, terminal conditions, risk events, snapshot
priorities, dataset admission criteria, required sensors, collection policy,
default review tolerance, admissible recovery patterns, and labels used by
dataset health.

\paragraph{Generation context.}
The task configuration is generated from the operator's task description, current camera
view, and robot/task context. The image matters: a visible drawer handle, an
occluded bin, or a deformable object can change phase definitions, visual hints,
and failure signals.

\paragraph{Artifact consumers.}
The same task contract is read by multiple downstream agents. VSA uses phases
and snapshot priorities to decide where to look online. The Post-Rollout Review Agent
uses terminal conditions, risk events, and admissible recovery patterns to
choose packet types and aggregation rules. The Data Governance Agent uses admission
criteria to route rollouts into training, review, rejection, or failure pools.
The Data Health Agent uses the task labels to report coverage and next-collection gaps.

\paragraph{Task-specific evidence.}
A placement task should over-sample release and final-settle events, while a
drawer task should emphasize contact onset, handle engagement, and motion after
contact. A manipulation task may define phases such as approach, grasp,
transfer, place, release, and final observation.

\paragraph{Phase vocabulary.}
The phase list is treated as a closed vocabulary for VSA, post-review, and
evaluation. Retry, correction, and intervention are represented as events or
failure/recovery evidence. This keeps phase statistics comparable across rollout
review, dataset health, and policy evaluation.

\paragraph{Training semantics.}
The Task Config Agent also specifies admission behavior. For example, recovered failures
can enter a separate failure pool while staying out of the primary imitation set
by default. The same rules are read by the VSA scheduler, post-rollout packet
builder, Data Governance Agent, and Master Agent.

\paragraph{Why explicit task rules matter.}
The same visual trace can mean different things across tasks. A brief contact
may be success evidence for drawer opening, a risk signal for cup placement, or
irrelevant for navigation. The Task Config Agent is where semantic perception becomes
task-grounded governance.

\subsection{Online Visual Snapshot Agent}
\label{app:vsa_agent}

The VSA has four internal submodules:
\begin{enumerate}[leftmargin=*]
    \item \textbf{Window Scheduler.} Converts event streams into bounded visual
    windows, coalesces redundant events, and protects critical windows.
    \item \textbf{Prompt Builder and Parser.} Constructs task-aware multimodal
    prompts and parses the response into phase, progress, risk, evidence, and
    confidence fields.
    \item \textbf{Phase State Machine.} Enforces monotonic phase progression by
    default, permits evidence-supported jumps, and handles retry after terminal
    failure.
    \item \textbf{Temporal Stabilizer.} Smooths jitter across adjacent windows
    and marks conflicts as review-worthy.
\end{enumerate}

\paragraph{Inputs.}
The VSA reads the robot profile, task config, rollout metadata, recent robot
events, raw image buffers, operator events, and previous stable phase state. It
can run before the full rollout finishes, creating sparse semantic anchors while
data collection is happening.

\paragraph{Outputs.}
Each snapshot artifact contains the window identifier, anchor time, event
reason, selected frame references, predicted phase, progress summary, risk
flags, visible evidence, confidence, review flag, parser status, and links back
to the rollout and task contract.

\paragraph{Runtime priority and permissions.}
Online VSA is prioritized during rollout collection. Offline review,
evaluation, and master summaries do not preempt the online worker. When a
rollout stops, raw capture closes first, then already queued VSA windows drain
before the rollout enters post-review. VSA only writes understanding artifacts:
phase, risk, uncertainty, and evidence references.

\paragraph{Lifecycle behavior.}
The VLM supplies visual perception. The agent maps that perception into phase
state, risk flags, and review-worthy uncertainty. This decomposition makes VSA
useful for downstream review while keeping final dataset decisions in the
offline governance path.

\paragraph{Failure handling.}
If the VLM times out, emits invalid JSON, or omits required fields, raw capture
continues. The agent writes a warning snapshot or review marker for the missing
semantics. Online perception remains helpful, and raw capture remains protected.

\paragraph{Details.}
Appendix~\ref{app:vsa} gives the event-windowing, contract-validation,
phase-stabilization, and final-observation details.

\subsection{Post-Rollout Review Agent}
\label{app:post_review_agent}

\paragraph{Role.}
The Post-Rollout Review Agent starts from an evidence index built over the raw rollout. It
combines online snapshots, terminal evidence, operator context, and raw
keyframes into a final annotation.

\paragraph{Packetization.}
The agent packages terminal evidence, post-terminal context, temporal context,
operator notes, and VSA snapshots into separate review packets. Packetization
lets the system weight different evidence sources and isolate failures.

\paragraph{Evidence index.}
The evidence index records raw video intervals, keyframes, VSA snapshots,
operator notes, stop reasons, synchronization warnings, missing streams, and
robot-log snippets. Packets are constructed from this index, giving each
downstream decision concrete evidence references.

\paragraph{Annotation schema.}
The final annotation contains a human-facing explanation and machine-facing
fields for dataset admission, dataset health, and future collection planning. It
records outcome, confidence, phase timeline, failure type, recovery status, risk
flags, key evidence references, packet-level warnings, and an L1 label.

\paragraph{Fallback behavior.}
The review agent sits downstream of both raw evidence and VSA snapshots.
Snapshots provide sparse semantic anchors. When an online VLM call is
unavailable or uncertain, the final annotation can still rely on raw keyframes,
operator notes, and terminal evidence.

\paragraph{Aggregation behavior.}
The aggregator treats final-state evidence as decisive for task success unless
post-terminal evidence contradicts it. Temporal packets explain how the rollout
reached the final state and where retries or failures occurred. Missing or
failed packets lower confidence and may route the rollout to human review, but
they do not erase the evidence produced by other packets.

\paragraph{VLM authority boundary.}
Offline VLM calls are allowed to provide terminal evidence, phase correction,
retry or failure evidence, label-quality hints, and training-usability hints.
Dataset admission is written later by governance rules. If a model response
contains an old or free-form dataset decision field, the post-review pipeline
treats it as advisory evidence and lets the Data Governance Agent write the formal
admission artifact.

\paragraph{Details.}
Appendix~\ref{app:review} details the evidence index, packet types, aggregation
rules, failure analysis, and admission semantics.

\subsection{Data Governance Agent}
\label{app:data_governance}

\paragraph{Role.}
The Data Governance Agent applies structured admission rules to the review
output. It is the lifecycle boundary between semantic interpretation and dataset
membership.

\paragraph{Inputs.}
The agent reads the final annotation, task admission policy, failure analysis,
operator notes, risk flags, confidence fields, and existing dataset state.

\paragraph{Admission classes.}
Typical admission classes are:
\begin{itemize}[leftmargin=*]
    \item \artifact{clean\_success}: clean success suitable for primary
    imitation training.
    \item \artifact{accepted\_with\_labels}: final success with useful recovered
    failure, retry, or phase labels.
    \item \artifact{successful\_but\_ambiguous}: trainable evidence that still
    deserves human inspection because labels or visual confidence are uncertain.
    \item \artifact{failure\_pool\_candidate}: useful failure evidence for
    diagnosis, recovery training, or targeted recollection, but excluded from
    the primary imitation set by default.
    \item \artifact{empty\_or\_insufficient}: unsafe, corrupted, off-task, or
    insufficiently evidenced rollout.
\end{itemize}

\paragraph{Separated fields.}
Semantic review and admission are kept as separate artifacts. The agent writes
separate fields for training eligibility, review priority, failure-pool routing,
exclusion reason, and operator override history. This separation lets a rollout
be useful for training while still requiring human inspection, or useful for
diagnosis while excluded from imitation training.

\paragraph{Decision versus training eligibility.}
The human-facing \artifact{decision} bucket can be \artifact{accepted},
\artifact{needs\_review}, \artifact{retry\_recommended}, or
\artifact{rejected}. Training selection uses the separate
\artifact{accepted\_for\_training} field, so a rollout can remain review-worthy
while still being eligible for training, and a human-approved failure can remain
outside the primary imitation set.

\paragraph{Rule order.}
The implementation treats safety and data integrity checks as hard exclusions,
then applies task-specific admission rules, then assigns review priority and
failure-pool routing. This order prevents a high-confidence visual success from
overriding corrupted streams, unsafe behavior, or missing terminal evidence.

\paragraph{Override semantics.}
Operator overrides are stored as lineage fields alongside the
automated decision. The artifact records who changed the decision, which field
changed, what evidence was cited, and whether the override affects training
eligibility, documentation review, or failure-pool membership.

\paragraph{Downstream use.}
The separated fields give the Data Health Agent enough structure to distinguish missing
coverage from poor execution, unsafe failure, ambiguous evidence, and label
uncertainty.

\paragraph{Lifecycle effect.}
The output is a dataset decision artifact. It links the review to accepted
rollout manifests, rejected or quarantined examples, failure pools, and later
dataset locks. Together, these links make data admission governable and
traceable across later training runs.

\subsection{Data Health Agent}
\label{app:dataset_health}

\paragraph{Role.}
The Data Health Agent turns rollout-level admission decisions into dataset-level
diagnosis. It assesses whether the current dataset is balanced, well evidenced,
and useful for the next training run.

\paragraph{Inputs.}
The agent reads accepted and rejected rollout artifacts, final annotations,
phase timelines, failure labels, confidence fields, missing-stream warnings,
operator interventions, and dataset-lock candidates.

\paragraph{Diagnostics.}
The health artifact reports phase coverage and the distribution of successes
and failures. It summarizes confidence and missing modalities, then checks
sample quality for duplication or corruption. Intervention rates and imbalance
help explain how representative the collection is. Useful failures remain
visible in the failure pool. These diagnostics distinguish a shortage of
successful demonstrations from a gap in a particular phase or weak sensor
evidence in an otherwise large dataset.

\paragraph{Computation.}
Health summaries use review and admission artifacts, including episodes outside
the primary training set. A small collection may lack an entire task phase.
A larger one may still be skewed or depend heavily on operator interventions.
The same records expose weak evidence and low-quality streams that simple
sample counts would miss.

\paragraph{Warnings.}
Repeated failures from one scene or an excess of easy successes can flag a
coverage problem. Missing terminal evidence or low confidence in a phase can
make its labels unreliable. Frequent operator intervention raises a separate
warning about the collection process. The agent also flags modality mismatch
with the target framework and possible train/eval leakage.

\paragraph{Lifecycle output.}
The Data Health Agent writes a training-readiness summary and a next-collection
recommendation. The Training Lifecycle runner reads this output before adapter
conversion, and the Master Agent uses it when proposing what the operator should
collect next.
Coverage counts, phase histograms, failure distributions, and distribution gaps
are computed deterministically; the Data Health Agent and Master Agent turn
those evidence fields into structured next-collection briefs, which are still
written as artifacts.

\paragraph{Why it is separate from admission.}
Admission decides whether individual rollouts may enter a dataset or failure
pool. The Data Health Agent decides whether the resulting collection is useful as a
training distribution. Keeping these agents separate prevents clean individual
examples from being mistaken for a complete dataset.

\subsection{Framework Discovery Agent and Dataset Adapter Agent}
\label{app:framework_adapter}

\paragraph{Framework Discovery Agent.}
The Framework Discovery Agent reads a training repository as a lab would use it: through
configuration files, dataset loaders, command-line entry points, expected data
layouts, and the commands already trusted by the team. The user's dataset,
training, and evaluation commands are treated as authoritative hints, so the host
repository can keep its own API.

\paragraph{Discovery output.}
The agent writes a target data contract, adapter-registry summary, candidate
command profile, output expectations, discovery event log, and integration
manifest. Optional code inspection helps infer loaders, checkpoint paths,
TensorBoard or log sources, and evaluation-result files.

\paragraph{Conservative inference.}
When multiple loaders, commands, or checkpoint locations are plausible, the
agent records alternatives and asks for confirmation through the integration
manifest. Explicit user commands outrank speculative repository interpretation;
the system adapts to the lab's working practice while preserving its training
conventions.

\paragraph{Dataset Adapter Agent.}
The Dataset Adapter Agent materializes data for the target framework by creating
manifests, symlinks, metadata views, or converted files. Raw robot data remains
read-only, and every converted output points back to accepted rollout
identifiers, admission artifacts, and hashes.

\paragraph{Adapter report.}
The adapter report records input rollouts, source hashes, output paths,
conversion strategy, dropped or remapped fields, camera/action naming, modality
availability, and warnings. A later policy can therefore be traced to the
governed dataset and to the exact data view presented to the learner.

\paragraph{Lineage behavior.}
The discovery result contains candidate train and evaluation commands, dataset
root assumptions, checkpoint output patterns, metric-log sources, adapter
confidence, and warnings for missing or ambiguous conventions. The adapter
records whether it used a direct manifest, a user-provided conversion command,
or a generated framework-specific view. The adapted data can be regenerated
from the dataset lock and adapter profile, keeping its relation to review
evidence explicit.

\paragraph{Framework coverage.}
The same contract pattern supports ACT-style behavior cloning, diffusion-policy
training, VLA fine-tuning, and LeRobot-compatible workflows. The governance
layer records what the external learner expects while leaving the learner intact.

\subsection{Training Monitor Agent and Version Governance Agent}
\label{app:training_monitor}

\paragraph{Training lifecycle.}
The Training Lifecycle runner builds a train manifest from post-review
admission artifacts, invokes the Data Health Agent, creates a dataset lock, adapts the
selected rollouts, launches the framework command, and records the resulting
training status.

\paragraph{Dataset lock.}
The dataset lock freezes accepted rollout identifiers, task and robot profiles,
adapter profile, artifact hashes, and the command context used to create the
training view. It is written before training begins, giving each policy
candidate a fixed dataset state.

\paragraph{Monitor output.}
The Training Monitor Agent records process state, command line, working directory,
environment hints, stdout and stderr summaries, parsed scalar metrics,
checkpoint candidates, evaluation outputs, warnings, and terminal error
categories using the profile discovered for that repository.

\paragraph{Failure handling.}
If training fails, the monitor still writes a status artifact with command,
environment hints, logs, error category, and partial outputs. Failed runs remain
part of lineage because they often explain why a candidate policy was not
produced or why a dataset needed adaptation.

\paragraph{Version governance.}
The Version Governance Agent creates immutable dataset locks and policy metadata. The
selected checkpoint is bound to its parent policy, dataset lock, framework
profile, adapter version, command hash, code revision, and observed metrics.

\paragraph{Policy metadata.}
Policy metadata records checkpoint hash, parent policy, compatibility
constraints, rollback target, training run identifier, dataset lock, framework
profile, metrics, and evaluation links. The reference implementation marks
policy metadata immutable after creation, protecting the identity of a policy
version once evaluation begins.

\paragraph{Why it matters.}
The lock-before-train discipline makes external learners auditable. If a policy
later behaves unexpectedly, the team can reconstruct the rollouts, reviews,
adapter, command, metrics, and checkpoint that produced it, even when the
learner itself is an unmodified repository.

\subsection{Policy Evaluation Agent and Deployment Governance Agent}
\label{app:eval_deploy_agents}

\paragraph{Evaluation.}
The Policy Evaluation Agent uses the same semantic evidence pipeline as post-rollout
review, with evaluation-specific criteria and parent-policy comparison. It
records per-trial success, failure phase, safety issue, regression against the
parent policy, evidence links, and weak task phases.

\paragraph{Evaluation context.}
Evaluation trials are linked to the candidate policy, parent policy, dataset
lock, task contract, and robot profile. This context separates broad policy
regression from failures that occur in task phases already known to be
under-covered in the training data.

\paragraph{Deployment governance.}
The Deployment Governance Agent reads evaluation artifacts together with policy metadata,
dataset health, coverage summaries, rollback candidates, policy hash, dataset
lock hash, and approval ticket. It then writes the lifecycle route: deploy,
hold, rollback, or collect more.

\paragraph{Executor boundary.}
The deployment artifact is a governance recommendation and audit record. The
reference implementation records release decisions while leaving policy-server
switching, rollback commands, and safety interlocks to the lab's approved
executor. Deployment and rollback therefore remain operator-confirmed actions
with their own approval trail.

\paragraph{Decision fields.}
The deployment ticket records recommendation, confidence, supporting evidence,
blocking risks, rollback target, required approvals, coverage caveats, and
next-collection hints. Deploy, hold, rollback, and collect-more decisions all
share the same evidence-linked ticket format.

\paragraph{Ticket semantics.}
The deployment ticket contains a recommendation plus reasons and lineage
evidence. A team can distinguish ``hold because coverage is weak'' from ``hold
because the candidate regresses on a phase the parent already solved.''

\subsection{Master Agent and Lifecycle Summary Artifacts}
\label{app:master_review}

\paragraph{Inputs.}
The Master Agent provides iteration-level memory. Setup and task artifacts
establish the current context. Dataset decisions and health reports describe
data readiness, while framework discovery and training status describe learner
progress. Evaluation and deployment artifacts show what happened to the policy
candidate. The agent reads these summaries alongside its previous memory and
emits a lifecycle summary at manual or automatic checkpoints.
Artifact hashes link each input to its source record and dataset lock. Stage
names and task phases locate the current work. Routing counts and unresolved
review queues expose data blockers; evaluation regressions and deployment-ticket
status expose policy blockers. Previous memory supplies the history behind these
issues. Free-form notes remain evidence strings attached to typed fields.

\paragraph{Low-token contract.}
Every major stage exposes a compact summary for the Master Agent. The agent
uses these summaries by default, which keeps cross-iteration synthesis cheap,
responsive, and focused on the next collection decision.

\paragraph{Health boundary.}
The Master Agent summarizes lifecycle state, risks, and next-collection options.
The Health view remains the operational diagnostic surface for data sources, ROS
topic freshness, recorder status, VSA workers, post-review workers, and training
services. The Master Agent consumes health summaries and leaves low-level
service diagnosis to the Health view.

\paragraph{Memory update.}
The memory artifact records persistent lessons across iterations: recurring
failure modes, stable task assumptions, known weak phases, effective collection
strategies, and framework-specific integration notes. These summaries are kept
compact, allowing future iterations to reuse the lesson without rereading every
raw video or log.

\paragraph{Outputs.}
It writes four artifacts: \artifact{master.state.json},
\artifact{master.review.json}, \artifact{master.memory.jsonl}, and a typed
\artifact{next\_collection\_brief}. The brief contains a route
(\artifact{collect}, \artifact{review}, \artifact{hold}, \artifact{train},
\artifact{rollback}), target phase, collection focus, evidence references,
confidence, and blocking risks.

\paragraph{Synthesis procedure.}
The Master Agent follows the same procedure at each checkpoint. It first
collects candidate issues from the Data Health Agent, review packets, evaluation
summaries, deployment tickets, framework status, and previous memory. It then
deduplicates issues by task phase and evidence source, ranks them by current
task risks and lifecycle blockers, and asks the language model to explain
only the top conflicts that cannot be resolved by fixed rules alone. The final
JSON must cite the source artifacts that support the recommendation. If a
required field is missing, if the top candidates conflict without enough
evidence, or if the confidence falls below the task threshold, the action route
becomes \artifact{review} or \artifact{hold}.

\paragraph{Next-collection planning.}
Suggested actions are intentionally grounded in existing lineage artifacts. The
agent may recommend collecting more data for a weak phase, inspecting ambiguous
reviews, rerunning framework discovery, training from a refreshed dataset lock,
holding deployment, or rolling back to a parent policy. The summary view records
the recommendation and evidence for the operator.

\paragraph{Advisory role.}
The Master Agent keeps humans oriented and proposes targeted collection or
inspection. It answers iteration-level questions: what changed since the last
policy, which failure mode dominates, whether the current blocker is data,
training integration, evaluation coverage, or deployment risk, and what a
data-collection operator should collect or inspect next.

\subsection{Design Rationale Across Modules}
\label{app:rationale}

\paragraph{Design principles in detail.}
The five principles in Section~\ref{sec:overview} serve as operational
constraints for the system. \textit{Traceability} lets a team recover why a
checkpoint exists by following its evidence back to the robot and task.
\textit{Semantic coverage} retains the observations needed to explain a slip or
an unstable terminal state. It also preserves recovered failures that a scalar
success metric would hide. \textit{Agent autonomy with explicit state} lets
agents make semantic and integration judgments through inspectable artifacts.
\textit{Portability} places local interface assumptions in profiles and
contracts, with adapters providing the learner's data view.
\textit{Operator efficiency} reduces routine video review and preparation while
routing ambiguous evidence to humans.

\paragraph{Why agent governance?}
Agent governance records context-dependent judgments as executable and
inspectable lifecycle operations. Fixed rules are appropriate for thresholds, hashes, locks, and
schema checks. Semantic questions need richer context: whether a failure is
trainable, whether a repository expects a particular data contract, or whether a
policy regression reflects a dataset gap. \system places agents at these
semantic transitions, then records their outputs as typed artifacts with
provenance, uncertainty, and bounded downstream effects.

\paragraph{Why lifecycle and lineage?}
Robot policy improvement unfolds as a sequence of dependent transformations. A
rollout is reviewed; review produces admission; admission changes a dataset
version; the dataset version produces a training run; the training run produces
policy metadata; evaluation and deployment decisions use that metadata. Lineage
records these relations in one artifact chain. The resulting record turns the
question ``why does this policy exist?'' into a concrete trace.

\paragraph{Why Robot Onboarding Agent?}
Multi-robot adaptation requires an explicit account of what each robot observes
and controls, which topics or files
correspond to camera, state, action, gripper, contact, and health, and which
warnings should follow from missing streams. The Robot Onboarding Agent localizes these
differences in a profile, giving downstream review, training, and evaluation
modules the same contract across robot types.

\paragraph{Why Task Config Agent?}
The same physical trace can mean different things under different tasks. A
contact event can be success for a drawer task, risk for a placement task, and
irrelevant for a navigation task. The Task Config Agent makes success criteria, phases,
risk events, snapshot priorities, and admission rules explicit. Agents then
interpret the video through the task context.

\paragraph{Why packetized VSA/post-review?}
A single VLM answer over a video rarely carries enough structure for policy
iteration. A policy team needs to locate the failed segment and determine whether
the terminal state was stable. That evidence must support training eligibility
and the dataset route, then remain linked to the later checkpoint and next
collection brief. Packetized review attaches frame references and confidence
to these judgments, so later stages can inspect and reuse them.

\paragraph{Why Dataset Admission and Data Governance Agent?}
Success, trainability, review priority, and high-value failure are different
concepts. A successful but ambiguous rollout may be useful for training while
still requiring documentation review; a failed rollout may be excluded from
imitation learning but valuable for failure taxonomy and targeted recollection.
The Data Governance Agent records training eligibility, review priority, and
failure-pool membership in separate fields.

\paragraph{Why Data Health Agent?}
Correct individual rollouts do not imply a ready dataset. A collection of clean
episodes can still over-represent easy scenes or miss a critical phase.
The Data Health Agent checks these distribution gaps alongside evidence quality
and intervention patterns. Its diagnosis helps the operator choose what to
collect next.

\paragraph{Why Framework Discovery Agent?}
Real robot-learning labs already have training repositories. Requiring every
team to rewrite its training stack around a new system would defeat lightweight
adoption. The Framework Discovery Agent reads the existing repository, commands, dataset
loaders, configuration files, and output conventions to infer a target data
contract and integration manifest. This lets the governance layer adapt to ACT,
diffusion-policy, VLA fine-tuning, or other learners without making those
learners native to RoboLineage.

\paragraph{Why Dataset Adapter Agent?}
Conversion is where lineage is often lost. A rollout may become an HDF5 file,
manifest row, symlink tree, or framework-specific data view, and the connection
to review evidence can disappear. The Dataset Adapter Agent materializes the target
format while preserving rollout identifiers, admission artifacts, hashes, and
dataset-lock ancestry.

\paragraph{Why Training Lifecycle, Training Monitor Agent, and Version Governance Agent?}
Training is often the least inspectable step: a command runs, a checkpoint
appears, and the relation to data decisions is reconstructed later by memory.
The training lifecycle makes this external process part of lineage. The monitor
preserves the run's logs and results, including failures that explain why no
usable checkpoint appeared. Version artifacts identify the frozen data and
adapter used by the learner. They also retain the training command and parent
policy, so the new checkpoint has a traceable identity.

\paragraph{Why Policy Evaluation Agent?}
Success rate alone hides the structure of policy quality. A candidate may have
the same average success rate as its parent while introducing a safety
regression, failing in a new phase, or succeeding only on scenes already covered
by the dataset. The Policy Evaluation Agent records trial evidence, weak phases, failure
types, regressions, and links back to the training lineage.

\paragraph{Why Deployment Governance Agent?}
Deployment decisions depend on the failure profile as well as success rate. A high success rate
may be unacceptable if failures are unsafe or concentrated in critical scenes;
a lower success rate may be useful if it fixes an important failure mode and
requests specific recollection. The Deployment Governance Agent turns evaluation
evidence, coverage, regression, rollback target, and policy provenance into an
inspectable deploy/hold/rollback/collect-more recommendation.

\paragraph{Why the Master Agent?}
Local artifacts explain local decisions, but policy iteration also needs a
cross-iteration memory. The Master Agent reads compact lineage artifacts and
summarizes the current stage, blocking risks, dominant failure modes, and next
collection target. Its lifecycle summary makes the accumulated effect of review,
training, evaluation, and deployment legible to the operator.

\paragraph{Why a frontend?}
A command-line lineage tracker may be auditable but still miss the rhythm of
collection sessions. The frontend lets operators control rollout capture and
inspect the evidence prepared by agents. Once uncertain cases are resolved and
dataset decisions approved, the operator can launch training and follow the
policy through deployment review. Agents maintain the records behind these
actions, while humans operate the robot and confirm consequential decisions.

\paragraph{Why lightweight artifacts?}
The system is intended to be adopted by robot labs without becoming a monolithic
robotics platform. JSON, YAML, JSONL, hashes, and schema validation are easy to
inspect, version, copy with datasets, and consume from existing scripts. This
keeps \system as a portable governance layer around existing robot-learning
stacks.

\paragraph{Design summary.}
Robot policy iteration needs semantic judgment and durable state in the same
loop. \system uses agents to produce task-grounded interpretations and lineage
artifacts to carry auditable lifecycle state across iterations.

\section{Artifact Schemas and Lineage Graph}
\label{app:schemas}

\subsection{Lineage Graph Semantics}
\label{app:lineage_semantics}

The lineage graph is a typed, versioned provenance record over lifecycle
artifacts. It is stored through ordinary files and hashes; the implementation
does not require a specialized graph database. The record makes policy
iteration auditable by showing which artifacts were read, which evidence
supported a decision, and which lifecycle transition produced a dataset, policy,
evaluation, or deployment ticket.

The graph is append-oriented. Each review, dataset decision, dataset lock,
training run, and deployment ticket is added as a node with provenance edges to
its dependencies. Corrections, operator overrides, and reruns appear as new
artifacts that supersede or annotate earlier artifacts. Rollback and audit stay
possible with lightweight storage.

\subsection{Node Types}

The core node types are:
\begin{itemize}[leftmargin=*]
    \item \artifact{robot.profile}: supplied and validated robot interface,
    stream bindings, calibration hints, safety limits, and warnings;
    \item \artifact{task.config}: phases, success criteria, risks, snapshot
    priorities, and admission rules;
    \item \artifact{collection.context}: operator, teleoperation interface,
    scripted routine, policy-assisted protocol, and intervention policy used for
    collection;
    \item \artifact{policy.metadata}: parent policy, checkpoint hash, dataset
    lock, framework profile, metrics, rollback target, and compatibility
    constraints;
    \item \artifact{rollout.record}: raw data paths, hashes, operator events,
    stop reason, synchronization status, collection context, and optional parent
    policy;
    \item \artifact{snapshot.vsa}: event-windowed semantic observation linked
    to selected frames and rollout events;
    \item \artifact{annotation.final}: post-rollout outcome, phase timeline,
    failure analysis, confidence, risk flags, and L1 label;
    \item \artifact{dataset.decision}: training eligibility, review priority,
    failure-pool routing, exclusion reason, and override history;
    \item \artifact{dataset.manifest} and \artifact{dataset.lock}: accepted
    rollout set, artifact hashes, adapter context, and frozen training view;
    \item \artifact{framework.profile}: discovered training repository
    contract, commands, expected data layout, metrics source, and outputs;
    \item \artifact{training.status}: process state, command, logs, metrics,
    warnings, checkpoints, and terminal errors;
    \item \artifact{eval.summary}: per-trial outcomes, regressions, weak phases,
    safety issues, and evidence links;
    \item \artifact{deployment.ticket}: deploy, hold, rollback, or collect-more
    recommendation with supporting evidence and approvals;
    \item \artifact{master.review}: current lifecycle state, persistent memory,
    blocking risks, and next-collection recommendation.
\end{itemize}

\subsection{Edge Vocabulary}

The graph uses typed provenance edges. A rollout \artifact{uses} a robot
profile, task config, and collection context; it may also
\artifact{references} a parent policy when collection is policy-assisted. A
snapshot \artifact{observes} a rollout window and \artifact{references} selected
frames. A final annotation
\artifact{reviews} a rollout and \artifact{references} snapshots, packets, and
keyframes. A dataset decision \artifact{admits}, \artifact{rejects},
\artifact{quarantines}, or \artifact{routes\_to\_failure\_pool} an annotation.
A dataset manifest \artifact{contains} accepted rollout identifiers. A dataset
lock \artifact{freezes} a manifest, hashes, task config, robot profile, and
adapter context. A training status \artifact{runs\_on} a dataset lock and
\artifact{uses} a framework profile. A policy metadata artifact
\artifact{was\_trained\_from} a training run and \artifact{descends\_from} a
parent policy. An evaluation summary \artifact{evaluates} a policy. A
deployment ticket \artifact{decides} over an evaluation summary and policy
metadata. A master summary \artifact{summarizes} compact lineage state and
\artifact{recommends} a next lifecycle action.

\subsection{Cross-Iteration Example}

In a typical iteration, a collection context, robot profile, and task config
produce rollout nodes. Online VSA snapshots \artifact{observe}
event-conditioned windows inside each rollout. The Post-Rollout Review Agent
\artifact{reviews} the rollout using raw keyframes and snapshots,
producing an annotation. The Data Governance Agent \artifact{admits} or
\artifact{rejects} that annotation and may route it to a failure pool. The Data
Health Agent \artifact{summarizes} the accepted and rejected decisions. A
dataset lock
\artifact{freezes} the accepted set before training. A training run
\artifact{uses} the lock and framework profile, and policy metadata
\artifact{was\_trained\_from} the resulting checkpoint. Evaluation
\artifact{evaluates} the candidate policy, the Deployment Governance Agent
\artifact{decides} whether to deploy, hold, rollback, or collect more, and
the Master Agent \artifact{recommends} the next collection target. The next
iteration inherits policy metadata and master memory from the same lineage
chain.

\subsection{Artifact Identity and Versioning}

Each artifact has a stable identifier, type, creation time, producer, schema
version, content hash, and dependency list. The dependency list is the producing
agent's local view of the lineage graph. When an artifact is superseded, the new
artifact records the superseded identifier and the reason, such as operator
override, updated task config, rerun review, new adapter, or failed training
retry. Earlier decisions can be corrected without erasing the audit trail.

\subsection{Minimal Rollout Schema}

A rollout record contains:
\begin{itemize}[leftmargin=*]
    \item rollout identifier, session identifier, robot identifier, task
    identifier, collection context identifier, and optional parent policy
    identifier;
    \item raw data paths and hashes;
    \item time interval, collection context, operator events, and stop reason;
    \item stream availability and synchronization status;
    \item links to snapshots, final annotation, and dataset decision.
\end{itemize}

In the reference implementation, raw robot data is the source of truth and is
recorded through a raw manifest. For ROS-based systems this manifest points to
the rosbag2 directory, recorded topics, recorder status, domain identifier, and
raw-data hashes. Semantic artifacts such as VSA snapshots and post-review
annotations reference the raw record.

\subsection{Minimal Dataset Decision Schema}

A dataset decision record contains:
\begin{itemize}[leftmargin=*]
    \item decision bucket, training eligibility, review priority, and label
    quality;
    \item failure-pool routing, exclusion reason, and retry recommendation;
    \item source annotation identifier, task admission policy, and confidence
    fields;
    \item operator override history and links to evidence used in the override;
    \item downstream links to dataset manifest, dataset health, and dataset
    lock.
\end{itemize}

\subsection{Minimal Dataset Lock Schema}

A dataset lock record contains:
\begin{itemize}[leftmargin=*]
    \item lock identifier, dataset manifest identifier, and selected rollout
    identifiers;
    \item source artifact hashes for rollout records, annotations, dataset
    decisions, robot profile, and task config;
    \item adapter profile, target training contract, and conversion report;
    \item creation command, producer, schema version, and compatibility notes.
\end{itemize}

\subsection{Minimal Policy Schema}

A policy record contains:
\begin{itemize}[leftmargin=*]
    \item policy identifier, parent policy, training run, framework profile, and
    checkpoint hash;
    \item dataset lock hash and accepted rollout count;
    \item training metrics and configuration hash;
    \item evaluation summaries and deployment tickets;
    \item rollback target and compatibility constraints.
\end{itemize}

These schemas are intentionally small. They support lightweight adoption first,
with richer schemas layered on top as labs need them.

\section{Online Visual Snapshot Agent Details}
\label{app:vsa}

\subsection{Event-Driven Windowing}

Uniform frame sampling spends review capacity on uninformative frames. VSA
schedules windows around robot, policy, and operator events. The implementation
uses events including \artifact{sequence\_start},
\artifact{gripper\_close}, \artifact{gripper\_open},
\artifact{gripper\_burst}, \artifact{contact\_transition},
\artifact{still\_start}, \artifact{motion\_resume},
\artifact{periodic\_sample}, \artifact{heartbeat}, and
\artifact{final\_observation}. A gripper burst suggests grasp or release; a
contact transition suggests a potential success, failure, or risk point; a
phase marker suggests a task boundary; and a rollout stop suggests either
terminal success or operator intervention. Robot evidence is sparse and
structured, and decisive information usually appears near an action boundary,
contact change, stop event, or final-state check.

The scheduler coalesces nearby events into one window when they describe the
same moment, but preserves protected events such as sequence start and final
observation. Periodic and heartbeat windows provide coverage when no
high-information event occurs, while high-information events take priority
when the online stream is dense. Each scheduled window records its anchor frame,
selected keyframes, covered events, and reason codes, giving later review the
reason the agent looked at that moment.

\subsection{Prompting and Contract Validation}

For each scheduled window, the VSA constructs a task-aware prompt from the
robot profile, task config, recent events, selected keyframes, phase hints, and
failure signals. The VLM is asked for phase, progress, visible evidence,
risk level, imminent-failure hints, confidence, and a review flag. The response
is validated against the snapshot contract before it becomes lineage evidence.
Invalid JSON, timeout, or missing fields do not stop raw capture. They produce a
conservative warning snapshot or a \artifact{needs\_review} marker.
A fluent model answer becomes governance evidence only after it satisfies the
fields that downstream review, admission, and training agents know how to
consume.

\subsection{Raw Capture Decoupling}

Raw capture is independent of semantic assessment. The raw log or rosbag is the
source of truth for training adapters and audit. VSA snapshots are sparse
interpretations attached to raw timestamps and frame identifiers. This
decoupling is important operationally: if the model is slow, unavailable, or
uncertain, the rollout is still captured, and the lifecycle records that the
semantic evidence is incomplete.
Semantic automation improves the lifecycle while raw evidence remains protected
from model latency or model failure.

\subsection{Phase Stabilization}

The phase state machine keeps online semantics stable. Let $p_t$ be the raw VLM
phase prediction at window $t$ and $\hat{p}_t$ be the stabilized phase. By
default, $\hat{p}_t$ cannot move backward relative to $\hat{p}_{t-1}$. Sparse
online evidence still needs to support real progress, so forward jumps are
allowed when the window has transition evidence, high visual confidence,
repeated phase observations, boundary events, or terminal evidence. Terminal
skips are treated more carefully: a jump to the terminal phase requires stronger
evidence unless risk is already high.

Rollback or retry requires explicit counter-evidence, such as a high-risk
window, imminent-failure flag, progress regression, gripper burst,
contact-transition event, or repeated visual evidence that the object or robot
returned to an earlier phase. This distinction is critical after terminal
actions: a real failure after release should be captured, while a single
ambiguous frame should not rewrite a successful lifecycle.

The temporal stabilizer complements the phase state machine. It smooths phase
and progress over adjacent windows, but treats risk asymmetrically: risk can
increase immediately, while decreasing risk requires consecutive confirming
windows. Boundary events such as gripper open, gripper close, motion resume, or
still start reset relevant history so old votes do not block real phase changes.
When the model is low-confidence or internally inconsistent, the stabilizer
marks the window as \artifact{needs\_review}.
The stabilizer gives downstream dataset decisions lifecycle-consistent phase
evidence despite jitter in online VLM predictions.

\subsection{Final Observation Handling}

Many robot tasks fail after the final action command. An object may slide after
release, a drawer may rebound, or a placed item may remain unstable. \system
records final observation as a dedicated snapshot. If a release event occurs,
the VSA waits for a settle interval. If the rollout ends without a
clean release, the rollout-stop event schedules a final window. Post-rollout
review then weights final evidence heavily. In the implementation these are two
separate paths: a release-settle path for nominal terminal actions and a
rollout-stop path for interrupted or ambiguous episodes. Both paths produce
lineage evidence for the final task state, which prevents a common failure mode
where the policy stops moving before the most important success evidence is
recorded.

\section{Post-Rollout Review Agent Details}
\label{app:review}

\subsection{Evidence Index}

The Post-Rollout Review Agent starts from an evidence bundle built over the raw
video. The bundle contains the task config, online VSA snapshots, window
records, selected keyframes, raw-frame references, operator notes, logs, and
final-observation coverage. The evidence index records frame spans, online phase
sequence, risk and progress summaries, visual disagreements, and source paths
used by review. Every final label can be traced back to the windows, frames, and
events that supported it.

\subsection{Evidence Packet Types}

The review pipeline uses four packet types:
\begin{itemize}[leftmargin=*]
    \item \textbf{Terminal focus:} final observation, terminal keyframes, task
    success criteria, and relevant VSA terminal claims.
    \item \textbf{Post-terminal context:} frames after terminal action or stop,
    used to catch unstable final states.
    \item \textbf{Temporal context:} selected windows around phase changes,
    risks, retries, and interventions.
    \item \textbf{Operator and log context:} stop reason, operator note, robot
    errors, missing streams, and synchronization warnings.
\end{itemize}
Final success, post-terminal
stability, failure localization, retries, operator context, and training
usefulness depend on different evidence. Packetization also provides failure
isolation: if one packet fails, the remaining packets can still support a
lower-confidence review, and the missing evidence is recorded explicitly.

\subsection{Aggregation Rules}

The aggregator treats terminal focus as decisive unless contradicted by
post-terminal evidence. Temporal context can identify failure phase and
recovery, but it cannot by itself declare task success if final state evidence
is missing. Missing packets lower confidence and may route to human review. The
aggregator also records reason codes so dataset decisions remain inspectable.
These rules keep success decisions grounded in terminal evidence while
preserving the history needed to diagnose failures.

Packet failures are isolated. If one VLM call times out or returns invalid
output, the remaining packets still contribute evidence. The failed packet is
recorded with its purpose, image frames, error type, and error message; the
rollout is marked for review when the missing packet could affect the decision.
Admission remains pending when a failed packet leaves essential evidence missing.

\subsection{Failure Analysis and Dataset Admission}

Failure Analysis reads the final phase timeline, online risk evidence, retry
events, and packet review outputs. It produces candidate failure segments,
failure type, severity, recovery status, high-risk frames, and final-failure
labels. The Data Governance Agent then separates rollout usefulness from documentation
uncertainty. Clean successes enter the primary training set. Successful but
ambiguous rollouts can be \artifact{accepted\_for\_training} while still routed
to review. Recovered failures are preserved for recovery training and failure
taxonomy. Empty, unsafe, corrupted, or insufficient episodes are excluded from
training. This separation is central to the system: \artifact{needs\_review}
is a human-attention signal, not the same thing as rejection.

\subsection{L1 Annotation}

The L1 annotation is a lightweight label embedded in the final annotation. It
contains outcome, task phase coverage, primary failure label, keyframes, and
training usefulness. Keeping L1 labels in the same final annotation avoids a
second manual labeling pass and ensures that dataset health, training selection,
and the review UI reference the same source record.

\section{Representative VSA and Post-Review Evidence Plates}
\label{app:vsa_review_plates}

\begin{figure}[p]
    \centering
    \includegraphics[height=0.39\textheight,keepaspectratio]{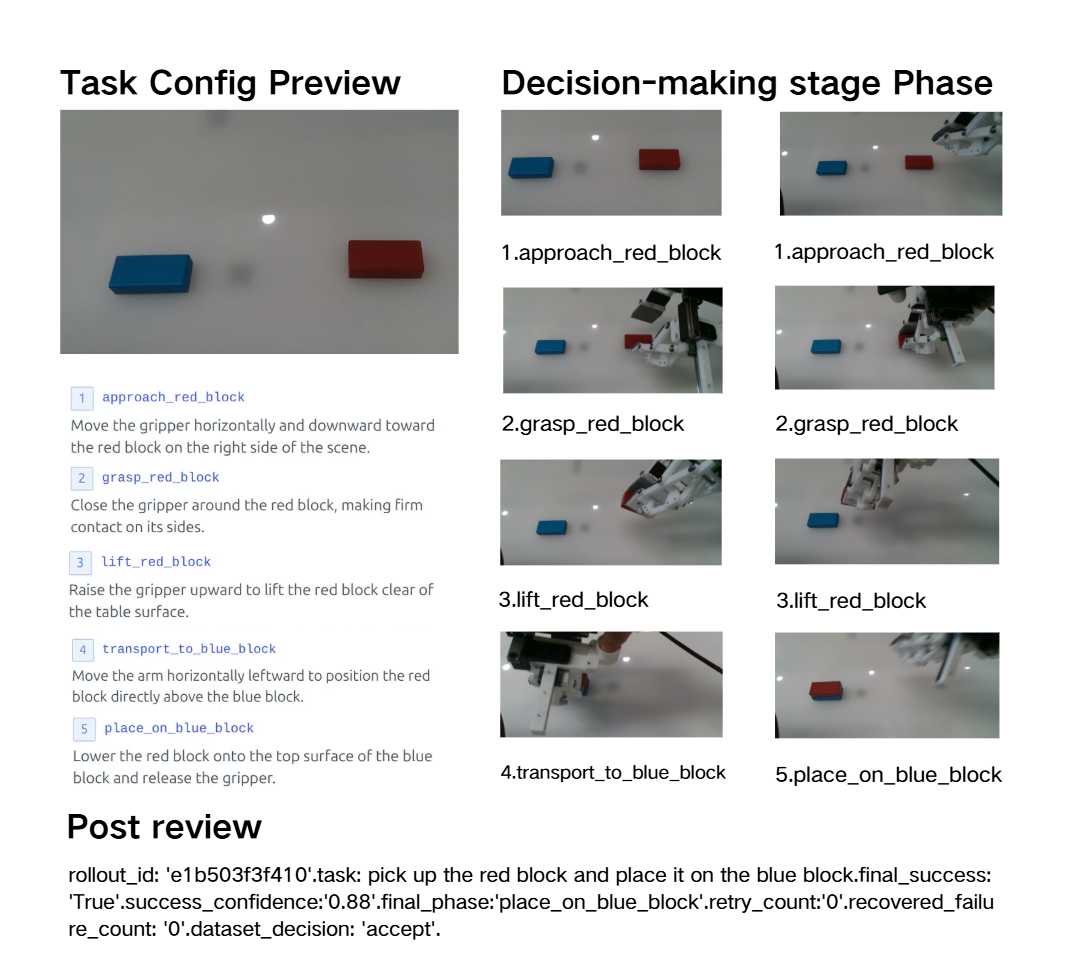}
    \vspace{0.8em}

    \includegraphics[height=0.39\textheight,keepaspectratio]{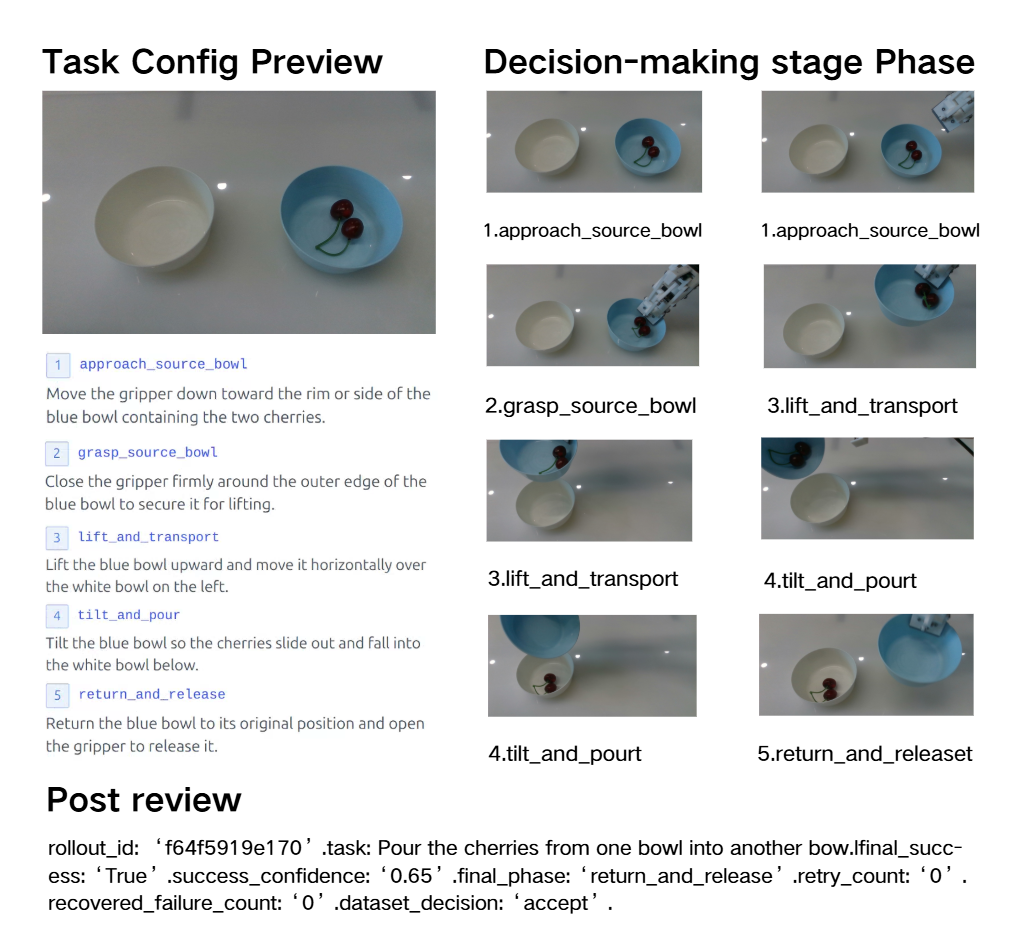}
    \caption{Representative VSA/post-review plates for standard pick-place and
    container-transfer tasks. The top plate shows an accepted block placement;
    the bottom plate shows a lower-confidence but accepted pouring rollout.}
    \label{fig:plate_arx}
\end{figure}

\begin{figure}[p]
    \centering
    \includegraphics[height=0.39\textheight,keepaspectratio]{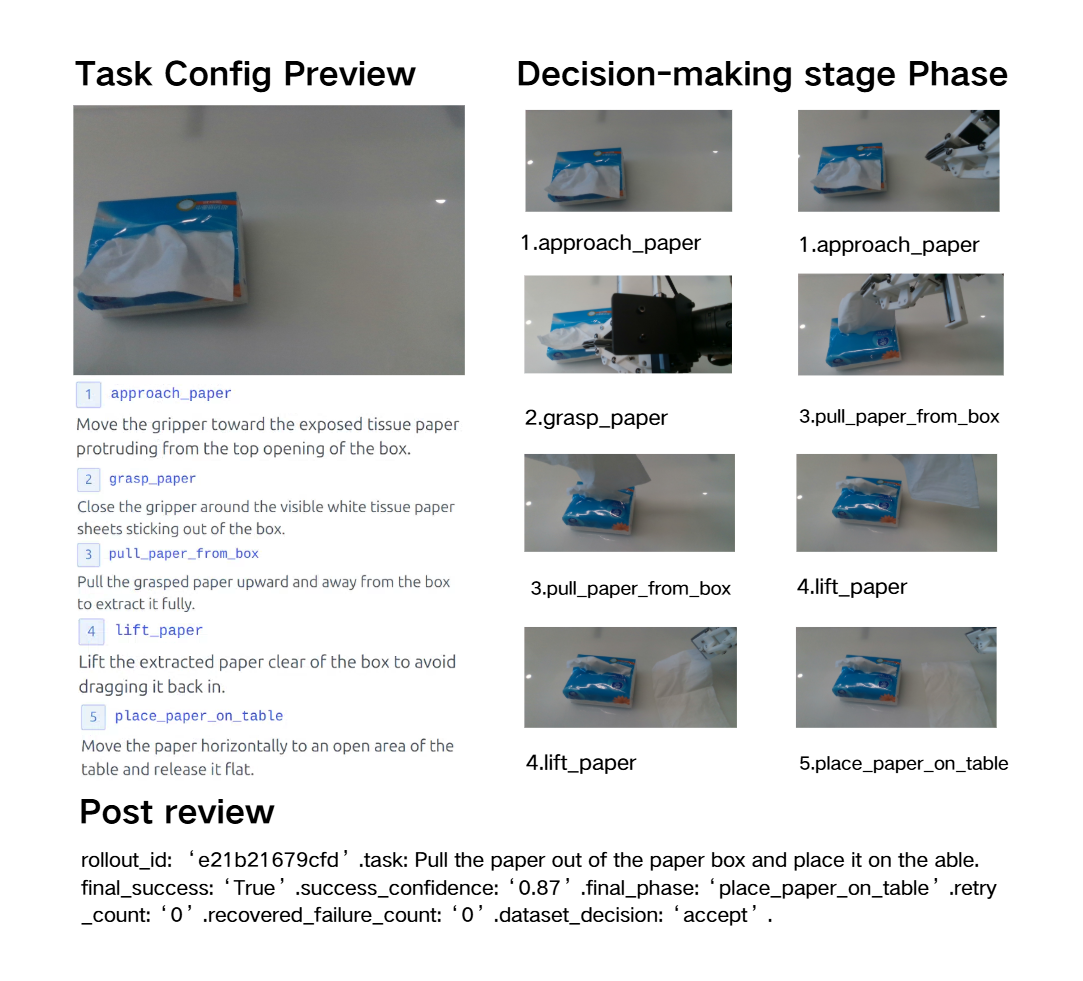}
    \vspace{0.8em}

    \includegraphics[height=0.39\textheight,keepaspectratio]{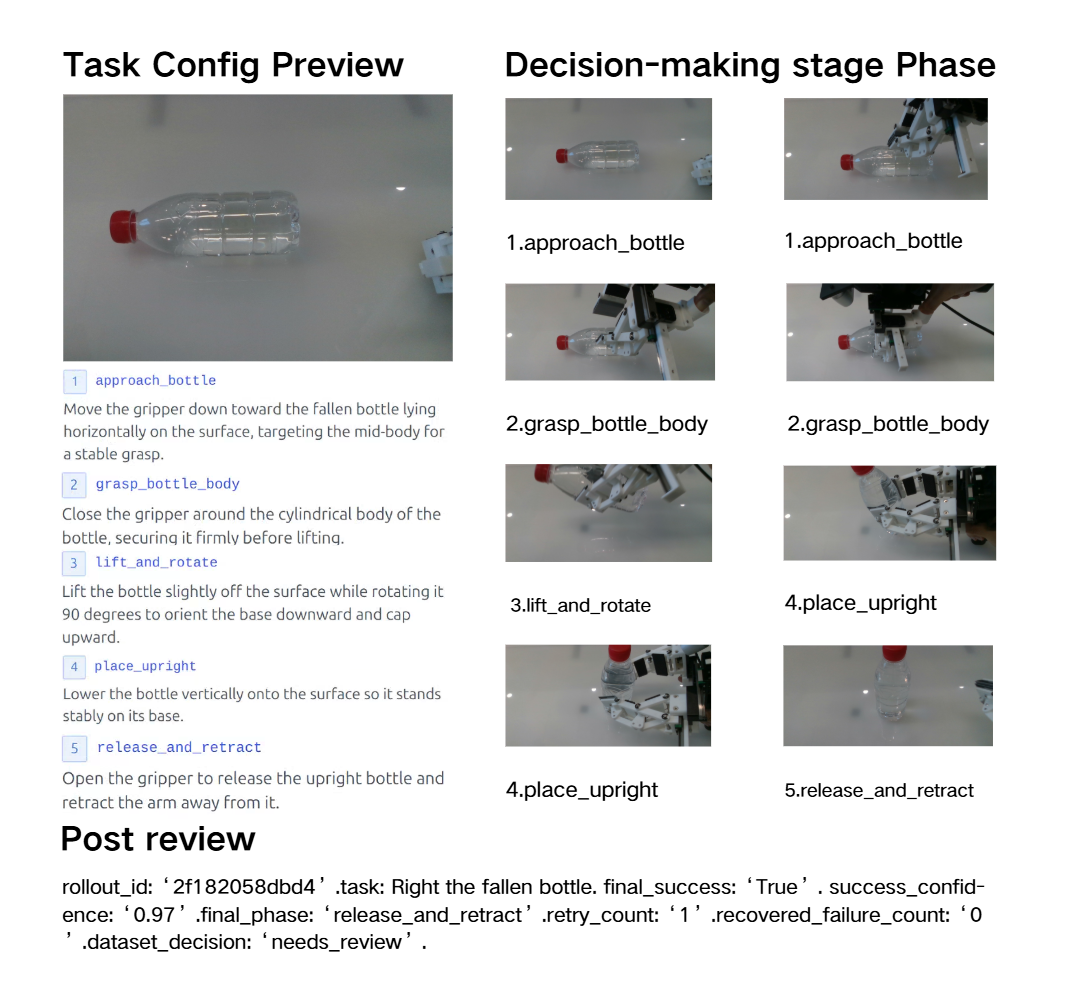}
    \caption{Representative VSA/post-review plates for thin-object extraction
    and object reorientation. These plates illustrate phase grounding beyond
    rigid pick-place and show how post-review records uncertainty or acceptance
    as a governed artifact.}
    \label{fig:plate_realman}
\end{figure}

\begin{figure}[p]
    \centering
    \includegraphics[height=0.39\textheight,keepaspectratio]{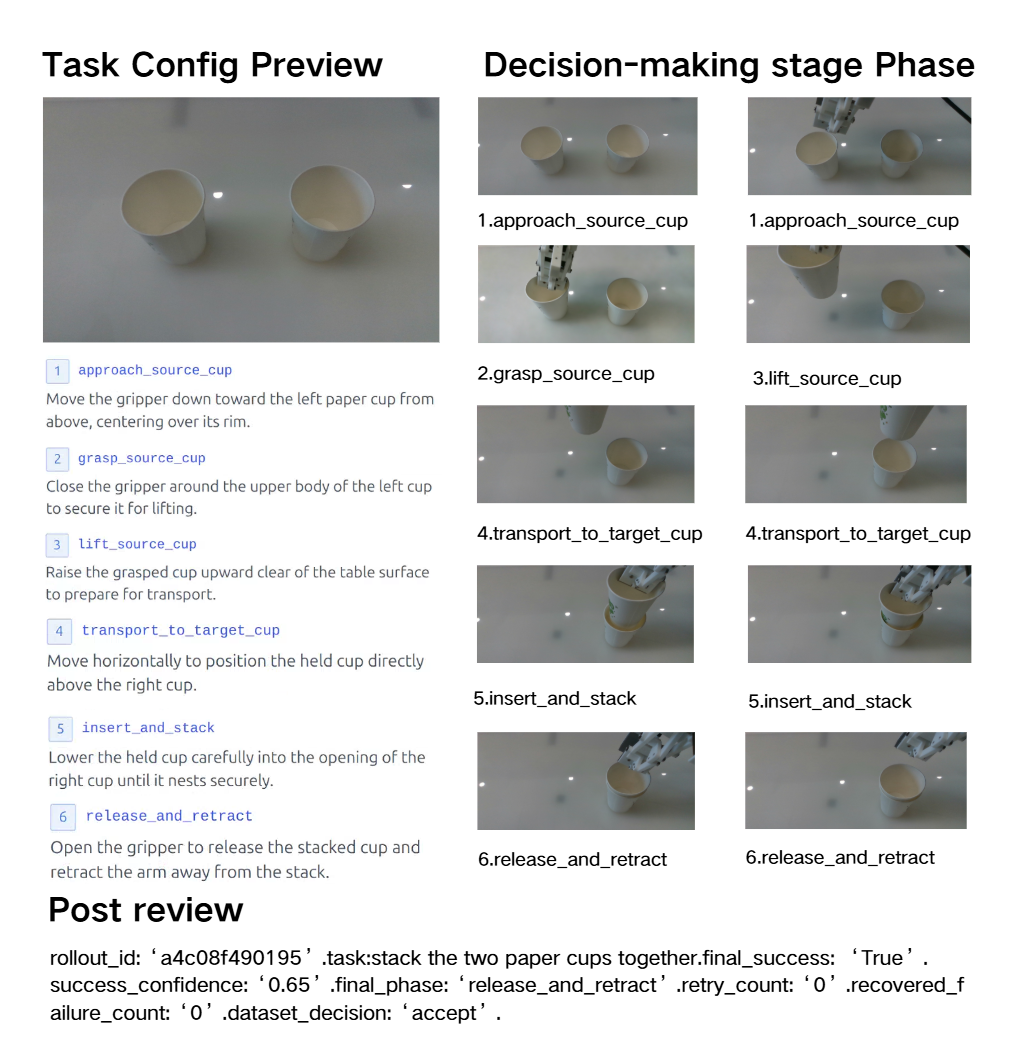}
    \vspace{0.8em}

    \includegraphics[height=0.39\textheight,keepaspectratio]{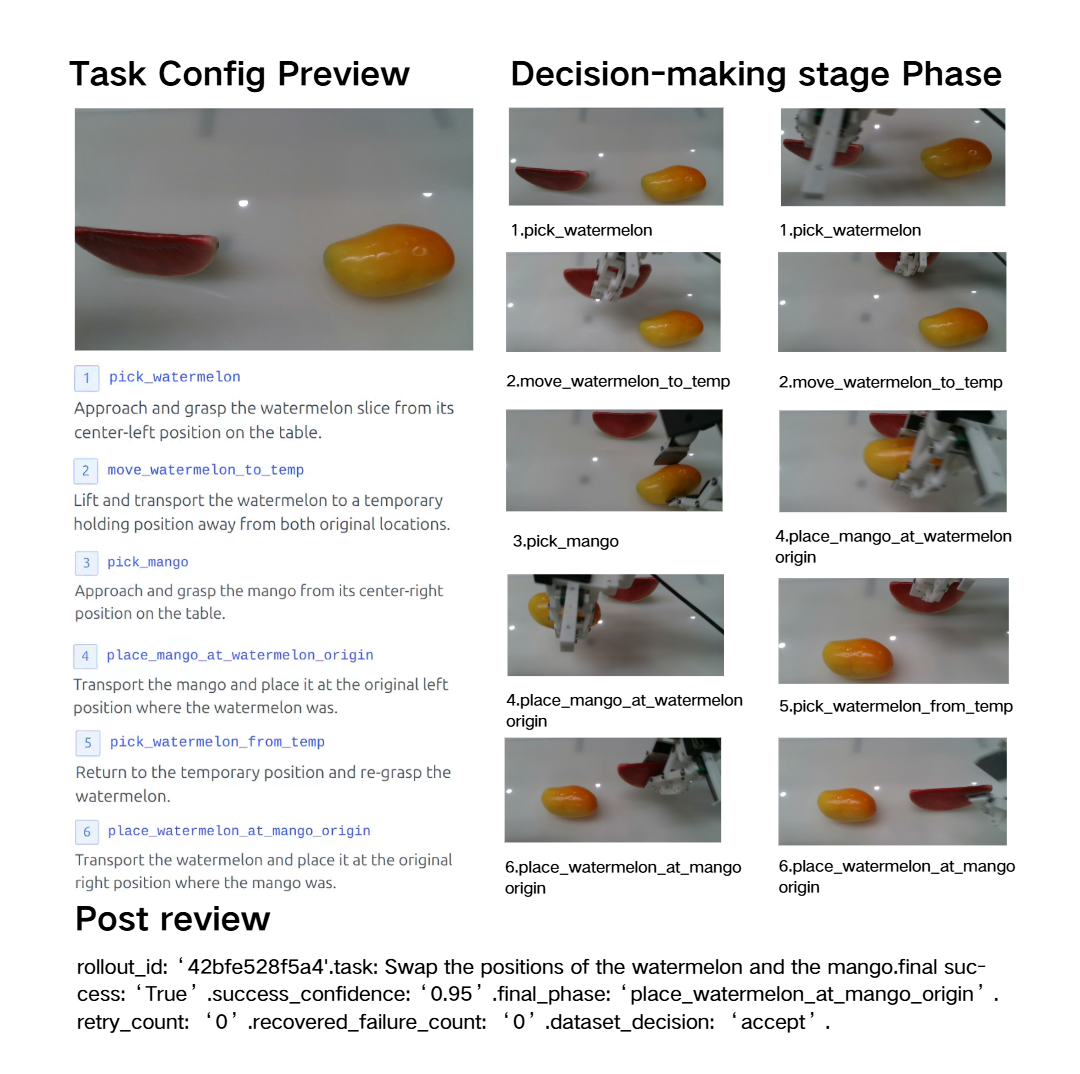}
    \caption{Representative VSA/post-review plates for precision insertion and
    long-horizon multi-object manipulation. The bottom plate preserves all six
    phases of the object-swap sequence, including intermediate object states.}
    \label{fig:plate_galbot}
\end{figure}

Figures~\ref{fig:plate_arx}--\ref{fig:plate_galbot} show six representative
VSA/post-review plates from the benchmark. Each plate combines the task
configuration preview, online VSA phase anchors, selected rollout frames, and
the post-review decision. Together they span success routing, review routing,
deformable and thin-object manipulation, reorientation, insertion, and
long-horizon multi-object manipulation.

\section{Training Integration Details}
\label{app:training}

\subsection{Discovery Signals}

The Framework Discovery Agent starts from command examples and README fragments.
It examines dataset loaders and configuration keys to infer the expected data
view. File names and imports provide additional clues, and previous framework
profiles supply reusable context. When multiple entry points remain plausible,
the agent proposes candidates for confirmation.

User-provided commands are authoritative hints because they encode the lab's
actual training practice. Repository structure and code snippets then explain
what those commands expect: dataset loaders reveal file formats, configuration
files reveal camera and action keys, and output conventions reveal where metrics
and checkpoints will appear.

The discovery artifact records the scanned file summary, inferred framework
type, user command intake, target data contract, selected adapter candidate,
output locations, warnings, and an event log. The integration manifest then
summarizes what later lifecycle stages need: training command, staging
directories, checkpoint glob, train-log path, evaluation result path, metrics
source, and dataset adapter strategy.

In our implementation, the same discovery interface covers ACT-style behavior
cloning, diffusion-policy training, VLA fine-tuning, and LeRobot-compatible
workflows. The discovered profile is descriptive: it records the repository's
existing expectations and leaves the learner's API intact.

\subsection{Target Data Contract}

The target data contract bridges governed robot rollouts and an external
learner. It records the expected dataset kind, input path template, camera
names, action/state assumptions, and adapter confidence. This contract lets
\system preserve one internal lineage graph while adapting to heterogeneous
training frameworks. Ambiguous contracts carry warnings and assumptions for
operator confirmation.

The contract is intentionally separate from the adapter implementation. The
contract states what the learner expects; the adapter materializes accepted
rollouts into that expectation. A team can replace an adapter or refine a
converter while preserving the lineage meaning of the dataset lock.

\subsection{Adapter Strategies}

The adapter chooses among three strategies:
\begin{itemize}[leftmargin=*]
    \item \textbf{Direct manifest:} the training framework reads the
    \system manifest directly.
    \item \textbf{User command:} a lab-provided command consumes the manifest
    and performs conversion.
    \item \textbf{Generated converter:} the adapter creates a framework-specific
    data view while preserving links to source rollouts.
\end{itemize}
All three strategies produce a dataset lock before training begins.

They also produce an adaptation report with input rollout identifiers, output
paths, file hashes, dropped or remapped fields, and warnings. This report
documents how a reviewed rollout becomes a training sample for the external
learner.

Raw robot data, review artifacts, and dataset locks remain authoritative.
Converted files are reproducible projections for a particular training
framework.

\subsection{Training Lifecycle Outputs}

Each training run has a workspace with its selected rollout manifest and dataset
health report. Adaptation status and framework results remain alongside the
training status. Policy metadata identifies the resulting checkpoint, and a
deployment recommendation records what should happen to it next.
The lifecycle context links the framework profile and dataset lock to the
supporting post-review summary. It then follows the run through evaluation and
deployment review. These links let an operator recover the data view and command
that produced a checkpoint, together with the recorded metrics and review
evidence.

The training status records where and when the command ran. Process state and
stdout/stderr summaries show execution progress; parsed metrics and checkpoint
candidates show training results. Warning codes and failure categories preserve
unsuccessful runs. Policy metadata fixes the selected checkpoint hash and parent
policy. It also identifies the dataset lock and framework profile, with the
adapter version and code revision needed to reconstruct the training view.
Evaluation links connect that policy to later trials.

\subsection{Policy Metadata Immutability}

Policy metadata is written after training and made immutable at the file level
in the reference implementation. The version check validates hashes whenever a
policy is evaluated or considered for deployment.

\section{Frontend Workflow}
\label{app:frontend}

The frontend turns lineage into a daily working interface. It exposes the
lifecycle through operator views for Robot Onboarding Agent, Rollout Control,
Review and Lifecycle, Training, Evaluation, Deployment Governance Agent, Master
Agent, and Data Health Agent.
The operator can:
\begin{itemize}[leftmargin=*]
    \item onboard a robot and inspect validated streams;
    \item define task phases, success criteria, risk events, and collection
    policy;
    \item run rollouts and watch online snapshots arrive;
    \item review final annotations with evidence packets and keyframes;
    \item approve, override, or comment on dataset decisions;
    \item inspect dataset health before training;
    \item launch framework adapters and monitor training;
    \item compare evaluation summaries across policies;
    \item inspect deployment tickets and rollback targets;
    \item read Master Agent summaries and next-collection plans.
\end{itemize}

Lifecycle governance helps only when it fits the rhythm of collection sessions.
An auditable command-line tracker is useful for records; the frontend makes the
same lifecycle actionable during repeated rollout and review cycles.

The frontend separates judgment from bookkeeping. Humans operate the robot,
inspect uncertain evidence, and approve consequential decisions; agents prepare
evidence, maintain artifact links, surface warnings, and propose the next
lifecycle action. This division of work is the practical mechanism behind the
operator-facing workflow evaluated in the main text.

\section{Experimental Protocol Details}
\label{app:protocol}

\begin{figure}[t]
    \centering
    \includegraphics[width=0.96\linewidth]{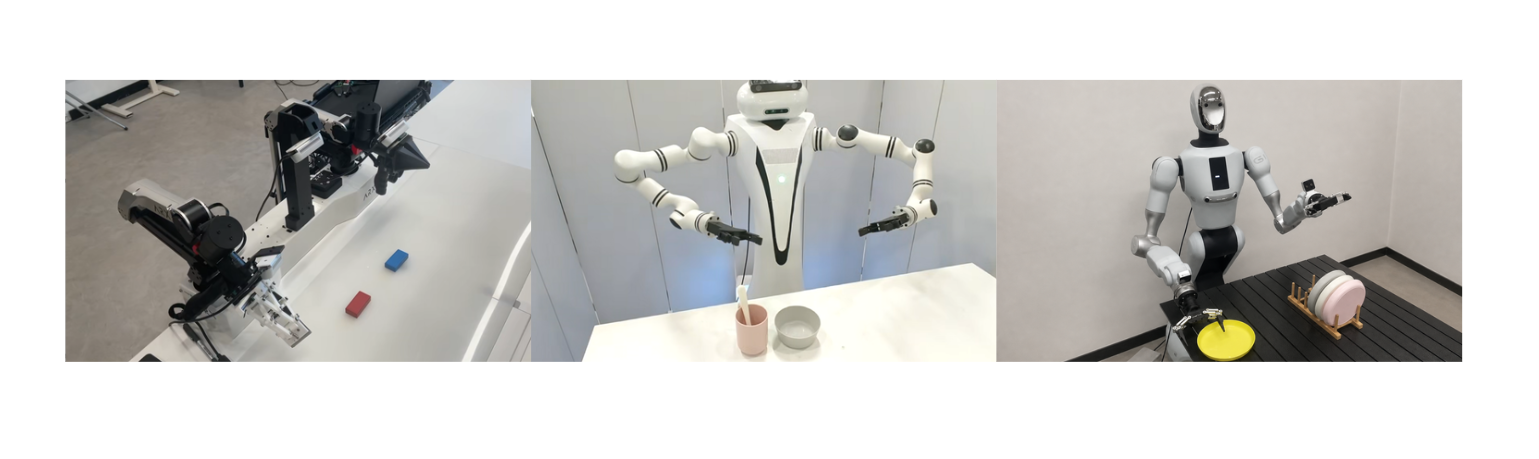}
    \caption{Robot platforms used in the evaluation. From left to right: ARX
    tabletop arm, Realman humanoid platform, and GALBOT G1 humanoid platform.}
    \label{fig:robots_app}
\end{figure}

\subsection{Annotation Protocol}

Expert annotators label rollouts using a common rubric. Each rollout receives:
task outcome, primary failure phase, secondary failure labels, safety or risk
flags, recovery status, and admission decision. Annotators see raw video and
robot logs but not \system's automated decision during ground-truth labeling.
Neither annotator authored the \system prompts, baseline definitions, or scoring
scripts. Final labels are fixed before model evaluation. Two experts
independently label the 500-rollout benchmark to estimate label consistency.
Before consensus adjudication, they agree on 97.0\% of outcomes,
91.0\% of primary failure phases, and 98.0\% of admission decisions; exact
agreement on the combined outcome/phase/admission tuple is 90.0\% with Cohen's
$\kappa=0.86$. The benchmark labels used for scoring are consensus labels
produced under the rubric. Outcome accuracy scores final success/failure,
failure-phase accuracy scores the primary failure phase, and admission accuracy
scores exact agreement with expert consensus over normalized data-use routes
across all 500 rollouts before operator overrides, under fixed admission rules.
Bad-data leakage is a stricter false-accept metric
over the 194 expert-rejected rollouts. Table~\ref{tab:semantic} reports point
estimates on this fixed 500-rollout benchmark.

Task contracts are frozen before benchmark scoring. For each task, the contract
is written from the natural-language goal, robot profile, available streams, and
two smoke rollouts used only to check phase names and success criteria. These
smoke rollouts are excluded from all reported metrics. After freezing the
contract, we do not edit prompts, phase definitions, admission rules, or
thresholds based on benchmark labels. Six late-added task variants were scored
after the prompt and schema set had already been frozen; they use the same
contract-generation procedure and serve as a small check against tuning to the
original task roster.

We stress-test task-contract brittleness with a 100-rollout replay.
Paraphrased task goals had little effect in pilot replays. Underspecified or
wrong success criteria reduced review quality and increased review routing.
Contract conflicts are recorded for inspection before primary-training admission.

\subsection{Baseline Definitions}

All semantic-review baselines use the same frozen Claude Sonnet 4.6 model
configuration and are scored against the same expert labels. The distinction is
what evidence is available before the final review artifact is written.
Table~\ref{tab:baseline_definitions} summarizes the controlled differences.

\begin{table}[h]
    \centering
    \scriptsize
    \begin{tabular}{p{0.30\linewidth}p{0.60\linewidth}}
        \toprule
        Method & Evidence available \\
        \midrule
        Direct video-to-VLM & Uniformly sampled frames or video summary; no VSA
        anchors \\
        VLM + schema only & Same frames as direct VLM, with schema validation
        and malformed-output retries \\
        VSA-only (no post-review) & Online VSA snapshots, final VSA memory, and
        task contract; no asynchronous packet review \\
        Post-review only (no VSA) & Raw keyframes and final observation; no online
        snapshot artifacts \\
        No packetized review & VSA snapshots and raw frames concatenated into a
        single review prompt \\
        No final observation / terminal focus & VSA and packet evidence except
        final settle/release window \\
        Full \system & VSA snapshots, packet evidence, final observation, and
        task contract \\
        \bottomrule
    \end{tabular}
    \caption{Controlled baseline definitions for Table~\ref{tab:semantic}.
    Baselines share the same model backend, rollout pool, task contracts, and
    scoring rubric; they differ in which evidence the VSA/post-review pipeline
    can use.}
    \label{tab:baseline_definitions}
\end{table}
The deterministic rule tree was written from the task-independent governance
rubric, schema fields, task contracts, and the two smoke rollouts used to check
each contract. It was finalized before benchmark scoring, did not use consensus
labels from the 500-rollout benchmark, and was not revised after inspecting
\system errors; authoring it took about 3.5 hours.

\begin{table}[h]
    \centering
    \scriptsize
    \begin{tabular}{lrrp{0.43\linewidth}}
        \toprule
        Diagnostic & Count & Rate & Meaning \\
        \midrule
        Primary training manifest & 296 & 59.2\% & Rollouts admitted into
        the training view after review and Data Governance Agent \\
        Routed to human review & 74 & 14.8\% & Ambiguous, low-confidence, or
        contract-conflicting cases held for inspection \\
        Failure pool & 76 & 15.2\% & Useful failures or recoverable traces
        preserved outside the primary training manifest \\
        Rejected / excluded & 54 & 10.8\% & Unsafe, wrong-task, corrupt, or
        irrecoverable rollouts excluded from training \\
        Operator override & 18 & 3.6\% & Human changes after inspecting
        routed evidence; stored as lineage fields \\
        False accept into primary manifest & 4 & 2.1\% of rejected &
        Expert-rejected rollouts that still entered primary training \\
        False reject / unnecessary exclusion & 14 & 4.6\% of accepted &
        Expert-accepted rollouts excluded or routed away from primary training \\
        \bottomrule
    \end{tabular}
    \caption{Routing and override diagnostics for the Full \system row in
    Table~\ref{tab:semantic}. The first four rows partition the 500-rollout
    benchmark before operator overrides; error rows use the same decisions.
    Overrides record later human changes within routed cases.
    False accepts use the 194 expert-rejected rollouts as denominator, and false
    rejects use the 306 expert-accepted rollouts as denominator.}
    \label{tab:routing_diagnostics}
\end{table}

\begin{table}[h]
    \centering
    \scriptsize
    \begin{tabular}{p{0.22\linewidth}p{0.34\linewidth}p{0.34\linewidth}}
        \toprule
        Error type & Representative case & Lineage containment \\
        \midrule
        False accept & A cube stack appeared stable in the terminal camera
        view, but the object slid outside the success tolerance after release &
        Post-terminal evidence was missing; the rollout was later linked to a
        terminal-settle rule update and marked as a reviewed exception \\
        False reject & A drawer-opening rollout stopped early because the
        operator released the handle after reaching the threshold; the visual
        packet looked like an incomplete pull & Operator override restored the
        rollout to the accepted set while preserving the original rejection
        reason and evidence links \\
        Contract mismatch & A long-horizon sorting rollout used a phase
        vocabulary from a pick-place-reset variant, making the intermediate
        success cue ambiguous & Data Governance Agent withheld admission and
        routed the rollout to review; the task contract was superseded by a new
        versioned artifact \\
        Model perception error & VSA confused visually similar red/blue blocks
        during approach, assigning the wrong target object to an online anchor &
        Packetized post-review corrected the final annotation from terminal and
        temporal evidence before admission \\
        Missing evidence & A Realman drawer rollout had usable motion logs but
        a dropped terminal frame & The rollout stayed outside the primary
        manifest and remained available as failure-pool evidence for diagnosis \\
        \bottomrule
    \end{tabular}
    \caption{Representative failure and containment cases for the Full \system
    row in Table~\ref{tab:semantic}. The records preserve the error, cited
    evidence, routing decision, and subsequent correction.}
    \label{tab:failure_examples}
\end{table}

\subsection{Policy Quality Protocol}
\label{app:policy_quality_protocol}

The human-comparison study uses eight tasks that are distinct from the two
closed-loop improvement case studies. For each task, we record a fixed
100-rollout candidate pool while \system's online VSA runs during collection.
The human expert workflow and the \system workflow then operate on the same raw
candidate pool and independently select 60 accepted demonstrations for training;
rejected and uncertain candidates remain outside the primary training set. Both
workflows use the same robot, task distribution, training family, training
command, and held-out 60-trial evaluation protocol. The human expert workflow
uses manual rollout review, manual dataset admission, spreadsheet or DVC-like
versioning, and hand-prepared training inputs. The \system workflow uses a
data-collection operator working through agent-generated evidence and framework
adapters. Figure~\ref{fig:quality_bar} therefore evaluates dataset admission and
workflow preparation from a shared candidate pool under matched training
budgets.

\subsection{Effort Study Protocol}
\label{app:effort_protocol}

For the effort study, operators process 12 matched 30-rollout pre-training
cycles, four per robot family, under manual and \system-assisted conditions.
Two trained data-collection operators run both workflows; condition order is
alternated by robot family, and the first practice cycle for each workflow is
discarded before timing. The manual workflow uses the lab's maintained scripts,
folder templates, review sheets, and version notes.
We count active human time before the training command begins: onboarding,
rollout collection, post-review/data admission, and training framework
integration. Physical rollout collection is counted equally in both workflows.
Manual sessions use the lab's normal video, spreadsheet, and data conversion
workflow. Operators inspect complete rollout videos, mark phases and
transitions, diagnose outcome and failure, record rationale, and assemble the
training manifest. Manual onboarding inspects robot configuration and topics
and edits recorder bindings. Manual framework preparation maps camera, state,
action, and timing fields, materializes files, edits configuration, and checks
training batches. \system sessions use the frontend, evidence packets, dataset
decisions, and automated adapter preparation. Human inspection of uncertain
rollouts, overrides, and admission approvals are counted in post-review time. We
count visible waiting or inspection time when it blocks the active user, but not
GPU training time, because both workflows call the same training stack once the
dataset is ready.
The reported 2-minute framework-integration cost is a per-iteration cost after
a framework profile and adapter have been created; one-time adapter authoring,
robot bring-up, and hardware debugging are not included in this routine-cycle
timing.
Online VSA overlaps raw collection, and asynchronous post-review overlaps later
rollouts. The measured 9.4-minute review stage is the residual blocking time for
unfinished or routed cases and manifest confirmation. Onboarding includes UI
task/topic entry followed by agent-generated and validated reusable profile
bindings. The validated profile is saved as YAML for reuse.

\begin{table}[h]
    \centering
    \scriptsize
    \begin{tabular}{lll}
        \toprule
        Item & Counted? & Scope \\
        \midrule
        Rollout collection, review, admission, adapter preparation & Yes &
        Repeated every cycle \\
        Robot profile / task contract validation & Yes, when used in-cycle &
        Reusable after validation \\
        Hardware bring-up, drivers, workspace construction & No &
        Lab-specific setup \\
        New adapter authoring or framework debugging & No &
        One-time integration \\
        GPU training time & No & Same command after dataset prep \\
        \bottomrule
    \end{tabular}
    \caption{Scope of the routine-cycle effort measurement.}
    \label{tab:effort_scope}
\end{table}

\begin{table}[h]
    \centering
    \scriptsize
    \begin{tabular}{lrrrrr}
        \toprule
        Workflow & Onboarding & Collection & Post-review & Framework prep. & Total \\
        \midrule
        Human expert workflow & $10.1{\pm}2.4$ & $40.5{\pm}4.8$ &
        $59.2{\pm}7.1$ & $20.4{\pm}3.0$ & $130.2{\pm}10.7$ \\
        \system workflow & $2.3{\pm}0.8$ & $40.6{\pm}4.6$ &
        $9.4{\pm}2.1$ & $2.0{\pm}0.7$ & $54.3{\pm}5.9$ \\
        \bottomrule
    \end{tabular}
    \caption{Active pre-training effort in minutes over 12 matched 30-rollout
    sessions. Values are mean $\pm$ standard deviation. Collection time is
    similar because the robot executes the same number of demonstrations; the
    savings come from review, admission bookkeeping, and framework preparation.}
    \label{tab:effort_variance}
\end{table}

\subsection{Storage Footprint}
\label{app:storage_footprint}

We account for storage over the same 500-rollout benchmark. Raw videos, robot
streams, and recording metadata average 72.6 MiB per rollout. Structured lineage
artifacts average 68 KiB per rollout, or 0.09\% of raw storage. The VSA keyframe cache
averages 16.3 MiB per rollout, or 22.5\%; it is a regenerable evidence cache.
Artifact generation is append-oriented, and summaries read indexed structured
artifacts. These measurements quantify the evaluated workload; sustained
latency and review backlog at larger scales remain future measurements.

\subsection{Model Backend Sensitivity and Retry Protocol}

All model-facing agents validate JSON outputs against their artifact schemas.
Invalid or unparsable responses are retried up to two times with the same
frozen prompt and model configuration. Online VSA timeouts, API errors, or
parse failures write conservative low-confidence snapshot artifacts and mark
the evidence for review; they do not block raw capture. Offline packet failures
are isolated to the failed packet, while successful packets still aggregate
into the final annotation. If missing evidence could affect admission, Data
Governance routes the rollout to review or excludes it from the primary
training manifest.

The main model route uses deterministic decoding where supported
(temperature zero and fixed prompt templates), fixed image sampling rules, and
logged request identifiers for replay. If a recurring failure cluster receives
contradictory or low-confidence semantic labels across rollouts, the Data Health Agent
opens a calibration warning and the Master Agent can recommend more inspection
or targeted recollection, but the cluster is not promoted into a clean dataset
decision until review resolves the contradiction. The warning and its resolution
remain linked to subsequent dataset decisions.

To separate the governance architecture from a single model backend, we replay
a stratified 200-rollout subset with Gemini 3.1 Pro Preview and qwen3-vl-plus
routes using the same task configs, packet prompts, schemas, and expert labels.
These routes are not tuned for the benchmark. Table~\ref{tab:model_sensitivity}
shows that semantic accuracy depends on model quality, while packetized
evidence, schema validation, and conservative admission still route uncertain
cases away from the primary training manifest.
For the full pipeline, weaker backends produce more review routes under the same governance rules,
increasing the inspection workload alongside the drop in semantic accuracy.

For the local open-weight replay, we use Qwen3-VL-32B-Instruct through vLLM on
two RTX 4090 GPUs modified to 48 GB each. The 200 rollout identifiers, expert
labels, prompts, task contracts, schemas, thresholds, and scoring are unchanged;
we do not retune the route. This local model is distinct from the hosted
qwen3-vl-plus route. It yields 165/200 correct outcomes, 146/200 correct phases,
167/200 matching admission routes, 6/100 false accepts, and 53/200 review routes.

\begin{table}[h]
    \centering
    \scriptsize
    \resizebox{\linewidth}{!}{%
    \begin{tabular}{lccccc}
        \toprule
        Route on 200-rollout replay & Outcome $\uparrow$ & Phase $\uparrow$ &
        Admission $\uparrow$ & Leakage $\downarrow$ &
        Review route \\
        \midrule
        Full \system, Claude Sonnet 4.6 & 95.5 & 87.5 & 96.0 & 2.0 & 13.5 \\
        Full \system, Gemini 3.1 Pro Preview & 94.5 & 86.5 & 95.0 & 3.0 & 14.5 \\
        Full \system, qwen3-vl-plus & 85.5 & 77.0 & 86.5 & 5.0 & 25.0 \\
        Direct video-to-VLM, Gemini 3.1 Pro Preview & 81.0 & 61.5 & 82.5 & 9.0 & 7.0 \\
        Direct video-to-VLM, qwen3-vl-plus & 72.0 & 52.0 & 75.0 & 14.0 & 7.5 \\
        \midrule
        Full \system, local Qwen3-VL-32B & 82.5 & 73.0 & 83.5 & 6.0 & 26.5 \\
        \bottomrule
    \end{tabular}
    }
    \caption{Model-backend sensitivity replay on a stratified 200-rollout
    subset. All values are percentages. Leakage is measured over 100
    expert-rejected rollouts in the subset.
    Gemini tracks the main backend closely. qwen3-vl-plus lowers semantic
    accuracy and increases review routing, but the governed VSA/post-review
    pipeline still keeps false admission below direct video review with the
    same backend. The local open-weight route uses Qwen3-VL-32B-Instruct.}
    \label{tab:model_sensitivity}
\end{table}

\subsection{Closed-Loop Recollection Protocol}
\label{app:closed_loop_protocol}

For each closed-loop task, we run three independent runs of three recollection
rounds. Before recollection, the initial policy $\pi_0$ is evaluated on 60 trials
in each run, for 180 trials per task. The reported initial success rates are
means over these three runs. Each round uses 30 accepted demonstrations for training and 30
diagnostic evaluation trials for failure analysis. The random baseline collects
additional rollouts using the same task distribution as the initial dataset,
so it measures the benefit of more data without failure-targeted recollection.
The human-expert condition asks an experienced robot-learning practitioner to
inspect the same diagnostic rollouts and propose the next collection focus. The
\system-guided condition uses Master Agent summaries, Data Health Agent artifacts,
and Deployment Governance Agent tickets to choose target phases, failure modes, and
scene variations. No lineage memory uses the same current-round VSA snapshots,
post-review packets, schemas, admission artifacts, and session-local coverage
summaries as \system, but forgets earlier rounds and policy ancestry. In
practice, this removes the persistent Data Health Agent state, parent-policy
outcomes, and prior next-collection briefs that connect a new failure to earlier
data gaps and policy changes. After the third round, all final policies are
evaluated on an independent 60-trial test set not used for recollection planning.

\section{Additional Results and Case Studies}
\label{app:additional_results}

\subsection{Task Roster and Qualitative Review Coverage}
\label{app:review_benchmark}

Table~\ref{tab:task_roster} lists the semantic-review benchmark at the task
family level. The benchmark intentionally mixes clean successes, recoverable
failures, wrong-object failures, phase mismatch, clutter, and long-horizon
partial completions so that reliable review cannot be obtained by checking only
the final image.

\begin{table}[h]
    \centering
    \scriptsize
    \setlength{\tabcolsep}{2.4pt}
    \begin{tabular}{p{0.18\linewidth}p{0.13\linewidth}p{0.48\linewidth}p{0.16\linewidth}}
        \toprule
        Family & Tasks / rollouts & Representative tasks & Notable review cases \\
        \midrule
        Blocks: rearrange / stack / obstacle / clutter &
        6 / 120 &
        pick up the red block and place it on the blue block; swap red/blue
        blocks; stack blocks on the blue block; place the blue block on the red
        block; move a cube around an obstacle; pick/place blocks under clutter &
        wrong-object grasp, OOD placement, obstacle placement, clutter-induced
        object confusion \\
        Containers and insertion &
        5 / 100 &
        pour cherries from one bowl into another; place a ball into a cup; place
        red/blue cubes into box or basket; place plush toys into left/right
        baskets; put all four blocks into a basket &
        partial completion, missing container, multi-object omission, unstable
        pour/transfer \\
        Contact-rich articulation &
        4 / 80 &
        pull a drawer and close it; pull a drawer, retrieve the red cube, and
        close it; open a box and pick up the cup inside &
        phase lag, partial handle engagement, task-config mismatch \\
        Deformable / thin objects &
        3 / 60 &
        pull toilet paper out and place it on the table; close an open book;
        stack two paper cups &
        visually subtle terminal success, failure to extract, incomplete close \\
        Orientation and tools &
        4 / 80 &
        turn a cup upside down; reset a fallen bottle; place knife and fork on
        a plate; hang scissors on a rack &
        high-risk orientation errors, one-object-only completion, drop events \\
        Long-horizon multi-object &
        3 / 60 &
        swap the positions of the watermelon and the mango; sort
        panda/elephant/tiger/lion plush toys into baskets; move toys in
        cluttered scenes &
        task over-decomposition, phase drift, multiple recoverable failures \\
        \bottomrule
    \end{tabular}
    \caption{Task coverage for the semantic-review benchmark. Table~\ref{tab:semantic}
    reports aggregate metrics and leakage over these families; qualitative
    examples below illustrate common VSA/post-review interactions. The 500
    rollouts were collected across 11 lab days, five workspace layouts, and 43
    physical object instances.}
    \label{tab:task_roster}
\end{table}

\begin{table}[h]
    \centering
    \scriptsize
    \begin{tabular}{p{0.36\linewidth}rrrrr}
        \toprule
        Task family & Rollouts & Outcome & Phase & Admission & Leakage \\
        \midrule
        Blocks: rearrange / stack / obstacle / clutter & 120 & 97.5 & 90.0 & 97.5 & 2.4 \\
        Containers and insertion & 100 & 97.0 & 91.0 & 99.0 & 0.0 \\
        Contact-rich articulation & 80 & 95.0 & 86.3 & 95.0 & 2.8 \\
        Deformable / thin objects & 60 & 93.3 & 83.3 & 93.3 & 3.6 \\
        Orientation and tools & 80 & 96.3 & 88.8 & 96.3 & 0.0 \\
        Long-horizon multi-object & 60 & 95.0 & 85.0 & 95.0 & 4.2 \\
        \midrule
        All & 500 & 96.0 & 88.0 & 96.4 & 2.1 \\
        \bottomrule
    \end{tabular}
    \caption{Per-family semantic-review diagnostics for the Full \system row in
    Table~\ref{tab:semantic}. Values are percentages. Leakage is the fraction
    of expert-rejected rollouts in each family admitted to primary training.}
    \label{tab:per_family_semantic}
\end{table}

\begin{table}[h]
    \centering
    \scriptsize
    \begin{tabular}{p{0.36\linewidth}rrrrr}
        \toprule
        Robot / policy family & Rollouts & Outcome & Phase & Admission & Leakage \\
        \midrule
        ARX / ACT-style action chunking & 200 & 96.5 & 88.5 & 96.5 & 1.8 \\
        Realman / Diffusion Policy & 180 & 95.0 & 86.7 & 95.6 & 2.6 \\
        GALBOT G1 / VLA/LeRobot-compatible workflow & 120 & 96.7 & 89.2 & 97.5 & 2.0 \\
        \bottomrule
    \end{tabular}
    \caption{Per-robot and per-policy-family semantic-review diagnostics for
    Full \system. The split matches the 25-task benchmark: 10 ARX tasks, 9
    Realman tasks, and 6 GALBOT G1 tasks. Values are percentages; leakage uses
    expert-rejected rollouts within each robot group as its denominator.}
    \label{tab:per_robot_semantic}
\end{table}

\paragraph{Qualitative review patterns.}
We observed four recurring interactions between VSA and post-rollout review.
First, online VSA sometimes lagged or confused visually similar objects, such
as red/blue block swaps, while post-review corrected the final outcome from
terminal and temporal packets. Second, when VSA detected high-risk events
during bottle reset, ball-in-cup, scissors, or drawer failures, post-review
usually routed the rollout to review while preserving it as trainable or
failure-pool evidence when appropriate. Third, task-config mismatch, such as a
drawer rollout evaluated under a bottle-reset phase vocabulary, surfaced as a
lineage inconsistency requiring inspection. Fourth,
long-horizon and cluttered tasks produced the hardest VSA errors, but
packetized review still preserved partial evidence and conservative routing.

\subsection{Closed-Loop Recollection Case Study}
\label{app:closed_loop_cases}

\begin{figure}[p]
    \centering
    \includegraphics[width=0.92\linewidth]{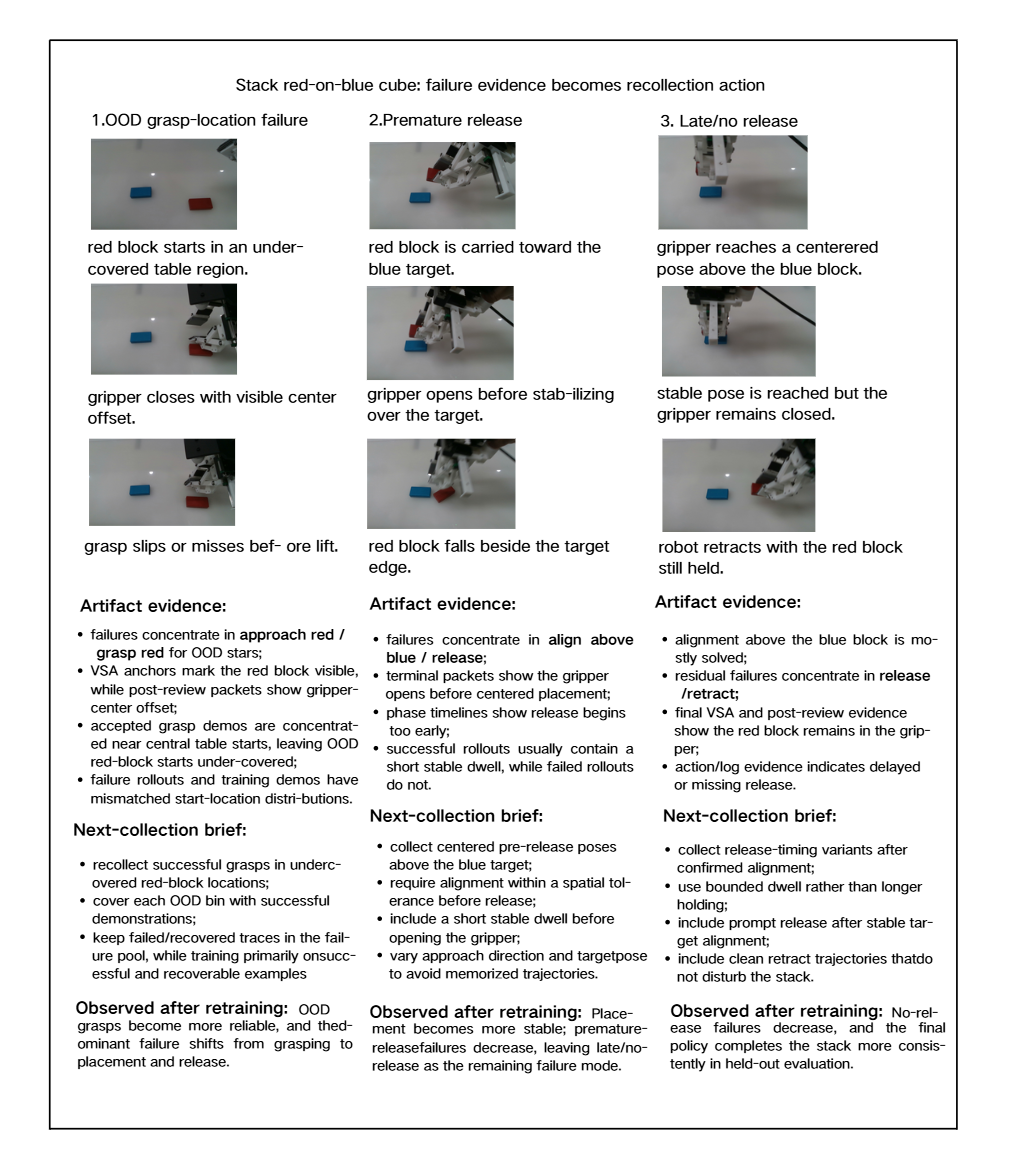}
    \caption{Stack red-on-blue closed-loop case. The plate follows three
    dominant failure stages through evidence artifacts, next-collection briefs,
    and observed changes after retraining.}
    \label{fig:stack_case_app}
\end{figure}

\paragraph{Case-study plate format.}
Figure~\ref{fig:stack_case_app} is an appendix plate for the stacking
closed-loop case. It uses three columns for the dominant failure stages that
emerge across successive policy iterations, and three rows for consecutive
frames from one representative failure. The text underneath each column is the
corresponding lineage artifact summary: observed evidence, next-collection
recommendation, and the observed result after retraining. Each frame reference
corresponds to a keyframe strip in the Appendix and release package, where the
same evidence packet can be inspected together with its artifact hashes.

The plate illustrates the mechanism behind the closed-loop results: lineage
artifacts make the current dominant failure explicit, turn it into a concrete
collection target, and link the observed changes after retraining to the
corresponding dataset update.

\section{Expanded Limitations and Future Work}
\label{app:future}

\paragraph{Evaluated scope.}
\system governs the data lifecycle around the lab's existing learner and
collection interface. Its decisions depend on the supplied task configuration
and the quality of robot streams. Evaluation criteria and safety thresholds also
come from the lab's configuration. Our experiments cover routine policy-iteration
cycles carried through this interface with agent-generated evidence and
governed artifacts.

\paragraph{Semantic uncertainty.}
\system can flag uncertainty and preserve the evidence behind a decision.
Its semantic interpretation can still be wrong, especially when the decisive
event is invisible to cameras. Force-dominant assembly, for example, may need
force or tactile measurements to establish success. Other criteria may depend
on sound or an external measurement. Our evaluation emphasizes tabletop and
visually grounded manipulation. Broader mobile manipulation and deformable-object
tasks need evaluation with evidence suited to their failure modes. Task-specific
sensor validation will also be needed where visual review is insufficient.

\paragraph{Model backend.}
The main evaluation uses one frozen Claude Sonnet 4.6 configuration. Appendix
Table~\ref{tab:model_sensitivity} reports hosted Gemini/Qwen and local open-weight
Qwen sensitivity replays; broader model tuning and prompt-perturbation studies
remain future work.

\paragraph{Governance beyond one lab.}
The current implementation is designed for a lab or small fleet. Sharing across
labs would require stable rollout identifiers and checks that another lab can
interpret the records. A federated version would also need to preserve lineage
when the underlying data stay private. Researchers could then share a dataset
with the evidence behind its admission decisions and the history of policies
trained and evaluated on it.

\paragraph{Richer correction and deployment-time collection modes.}
The evaluated workflow covers ordinary rollout capture and post-hoc review.
One extension is raw collection for generalization studies, without online
semantic review. Another is reference correction: a policy proposes actions for
inspection without driving the robot. Deployment monitoring could record when
an operator takes over and how the execution is corrected, preserving those
traces in lineage. Policy-foresight or AR trajectory displays could help the
operator inspect model intent while remaining separate from governance agents.
The artifact model can represent these modes, but their interfaces and safety
procedures still need validation.

\paragraph{Lifelong robot memory.}
\system currently summarizes each policy iteration and links it to previous
policies. A longer-lived memory could retain recurring failures and the data
that helped resolve them. It could also track sensor reliability and revisit
release decisions after deployment. These lessons should remain linked to
inspectable evidence and versioned records across iterations.

\paragraph{Learning governance decisions.}
Our dataset and deployment transition rules are explicit by design. Over time,
systems may learn better priors for which ambiguous rollouts are worth human
review, which failure pools improve robustness, or which evaluation regressions
predict deployment risk. The challenge is to learn these priors without
removing accountability from the final decision.

\paragraph{Benchmarks for data ecosystems.}
Robot learning benchmarks usually score policies. The workflow that produced
a policy also deserves evaluation. A benchmark could measure how reliably
episodes are reviewed and how quickly they become training-ready data. It could
test whether failure diagnosis leads to useful recollection. Complete lineage
would let evaluators trace those decisions and check whether the training run
can be reproduced.

\paragraph{Community interface standards.}
The field would benefit from simple standards for robot rollout manifests,
semantic review artifacts, dataset admission records, and policy metadata.
\system is one concrete proposal, but the deeper goal is interoperability. A
policy trained in one framework should be able to carry its data lineage into
another evaluation or deployment environment.



\end{document}